\documentclass[runningheads]{llncs}

\usepackage{amsmath,amsfonts,bm}

\def\1{\bm{1}}

\DeclareMathAlphabet{\mathsfit}{\encodingdefault}{\sfdefault}{m}{sl}
\SetMathAlphabet{\mathsfit}{bold}{\encodingdefault}{\sfdefault}{bx}{n}

\DeclareMathOperator*{\argmax}{arg\,max}

\usepackage{hyperref}

\usepackage{url}
\usepackage{mathtools}
\usepackage{amsmath}
\usepackage{todonotes}
\usepackage{subcaption}
\usepackage[inline,shortlabels]{enumitem}
\usepackage{pgfplots}
\usepackage{booktabs}
\usepackage{siunitx}
\usepackage{algpseudocode}
\usepackage{algorithm}

\usepackage{xparse}

\NewDocumentCommand{\Interp}{smO{\curIm}O{\basLn}}{
  \IfBooleanTF{#1}
  {x_{#2}}
  {[#3,#4]_{#2}}
}

\graphicspath{{./graphics/}}

\RequirePackage{aliascnt}

\newcommand*\dd{\mathrm{d}}

\newcommand*\one{\mathbf{1}}

\newcommand*\liealg[1]{\mathfrak{g}}
\newcommand*\RR{\mathbb{R}}
\newcommand*\vcr{\mathbf{r}}
\newcommand*\curIm{x_0}
\newcommand*\basLn{x_{1}}

\newcommand*\Vect{V}
\newcommand*\Mod{\mathcal{I}}
\newcommand*\Rng{\mathcal{M}}
\newcommand*\BoRng{\mathcal{M}_{\{0,1\}}}
\newcommand*\Cont{\mathcal{C}}
\newcommand*\Pth{\mathcal{P}}
\newcommand*\Score{P}
\newcommand*\ScoreRet{{\Score_{\uparrow}}}
\newcommand*\ScoreDiss{{\Score_{\downarrow}}}
\NewDocumentCommand\pth{o}{\varphi\IfValueT{#1}{_{#1}}}
\newcommand*\pthRet{\pth[\uparrow]}
\newcommand*\pthDiss{\pth[\downarrow]}
\newcommand*\ScoreCtr{{\Score_{\updownarrow}}}
\newcommand*\ScoreBndStr{{\Score_{\uparrow\downarrow}}}
\newcommand*\pthBndStr{\pth[\uparrow\downarrow]}
\newcommand*\Abl{\mathcal{A}}
\newcommand*\Lspace[1]{\mathcal{L}^{#1}}
\newcommand*\satStrength{{\zeta_\text{sat}}}
\newcommand*\pinchStrength{{\zeta_\text{pinch}}}

\newcommand*\linpath{\ell}

\DeclarePairedDelimiter\norm{\lVert}{\rVert}
\DeclarePairedDelimiter\paren{\lparen}{\rparen}
\DeclarePairedDelimiter\bracket{\lbrack}{\rbrack}

\DeclarePairedDelimiterX\pairing[2]{\langle}{\rangle}{#1,#2}
\DeclarePairedDelimiterX\scalprod[2]{\lparen}{\rparen}{#1,#2}
\DeclarePairedDelimiterX\set[1]{\{}{\}}{\,#1\,}
\DeclarePairedDelimiterX\setc[2]{\lbrace}{\rbrace}{\,#1 \;\delimsize\vert\; #2\,}

\begin{document}

\pagestyle{headings}
\mainmatter

\title{Ablation Path Saliency}

\institute{Western Norway University of Applied Sciences}
\author{Justus Sagem\"uller \inst{1}\orcidID{0000-0003-1882-1096} \and
 Olivier Verdier \inst{1}\orcidID{0000-0003-3699-6244}
\texttt{\{over,jsag\}@hvl.no}
}
\titlerunning{Ablation Path Saliency} 
\authorrunning{J. Sagem\"uller, O. Verdier} 

\maketitle

\begin{abstract}
  Various types of saliency methods have been proposed for explaining black-box classification.
  In image applications, this means highlighting the part of the image that is most relevant for the current decision.

  Unfortunately, the different methods may disagree and it can be hard to quantify how representative and faithful the explanation really is. 
  We observe however that several of these methods can be seen as edge cases of a single, more general procedure based on finding a particular \emph{path} through the classifier's domain. This offers additional geometric interpretation to the existing methods.
  
  We demonstrate furthermore that ablation paths can be directly used as a technique of its own right. This is able to compete with literature methods on existing benchmarks, while giving more fine-grained information and better opportunities for validation of the explanations' faithfulness. 



\end{abstract}

\section{Introduction}

The basic idea of \emph{saliency} or \emph{attribution}
is to provide insights as to why a neural network produces a given output (for instance, a classification) for a given input (for instance, an image).
There is no clear consensus in the literature as to what saliency should exactly be, but various properties that such a method should fulfill have been proposed.
All the methods discussed here start out by contrasting the given input (also called \emph{current target}) with another one, called \emph{baseline}, which should be neutral in at least the sense of not displaying any of what causes the target image's classification.
The saliency problem then amounts to finding out what the features of the target are which cause it to be classified differently from the baseline.

In \cite{DBLP:journals/corr/SundararajanTY17} the authors give axioms attempting to make it precise what that means. Of these, \emph{sensitivity} captures most of the notion of saliency,
 namely, that the features on which the output is most sensitive should be given a higher saliency value.
 The authors give further axioms to narrow it down: implementation invariance, completeness, linearity and symmetry preservation.
 They obtain a corresponding method: the {Integrated Gradient} method.
 Despite the attempt to thus narrow down the choice of saliency method, Integrated Gradient has not established itself as a default in the community.
 Indeed the axioms used to justify it are not altogether self-evident.

In \cite{fong17_inter_explan_black_boxes_by_meanin_pertur}, a method is provided whose construction is quite different.
Instead of following axioms about the properties a method should have, they produce a result that has direct meaning associated to it, namely as a mask that preserves only certain parts of the input and removes others, optimised so that the classification is retained even at high degrees of ablation, i.e., when the mask only keeps small part of the target image.
This method is highly appealing, but in practice the optimisation problem is ill-conditioned and can only be solved under help of regularization techniques. That prevents this technique, too, from being a definite saliency method or ``the'' saliency method.

Various other methods from the literature are in a broadly similar position, all with certain arguments for their use but also various practical limitations and no clear reason to favour them over the alternatives.
In some cases there are evident mathematical relationships between the methods, but they have not been investigated thoroughly yet or exploited for a unifying generalization.

This is what our paper provides: it introduces \emph{ablation paths}, which take up and extend the idea of integrating from the baseline to the target image. It combines this with the notion of ablation / masks, in that each step along the path can constitute a mask highlighting progressively smaller portions of the image.
The main purpuse of this is mathematical unification and better (meta-) understanding of the various methods, but ours can also be used as a saliency method by itself.

\begin{figure}[htb]
  \centering
    \includegraphics[width=\textwidth]{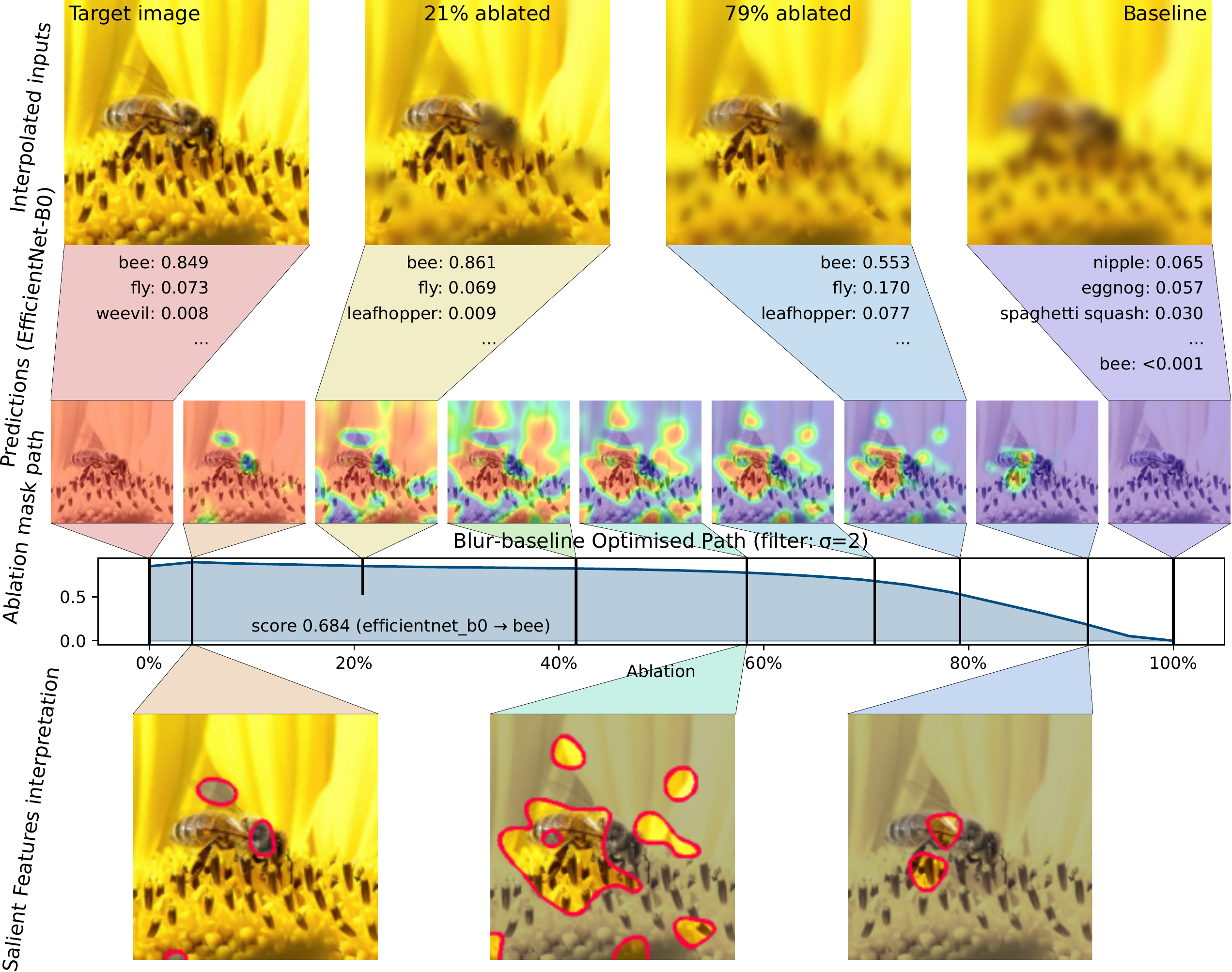}
\caption{%
Example of how an ablation path (sequence of masks, middle row) gives rise to a transition between a current target (a bee from ImageNet) and a baseline (blurred version of the same image). 
}
\label{fig:path-concept-overview}
\end{figure}

A summary how the method works:
suppose first that images are defined over a domain \(\Omega\), which can be regarded as the set of pixels in the discrete case, or as a domain such as a square, for the image at infinite resolution.
We define \emph{ablation paths} as parameter dependent smooth masks \(\pth \colon [0,1] \to \Cont(\Omega,\RR)\),
with the further requirement that the mask at zero, \(\pth(0)\), should be zero over the domain \(\Omega\), and the mask at one, \(\pth(1)\), should be one over the domain \(\Omega\).
We also impose that, at each pixel, the mask value increases over time (see \autoref{fig:path-concept-overview}),
and that this happens with a constant area speed: the area covered by the mask should increase linearly over time (see \autoref{sec:ablationpath}).
Let \(F\) be the classifier, which outputs a probability between zero and one.
We choose a current image of interest \(\curIm\) and a \emph{baseline image} \(\basLn\).
The objective function \(\ScoreRet\) is then \(\ScoreRet(\pth) := \int_{0}^1 F(\curIm + \pth(t)(\basLn-\curIm)) \,\dd t\) (see \autoref{sec:score}).
Assuming that \(F(\curIm) \simeq 1\)  and \(F(\basLn) \simeq 0\), maximising the objective function means that we try to find an ablation path that stays as long as possible in the decision region of \(\curIm\).
Intuitively, we try to replace as many pixels of \(\curIm\) by pixels of \(\basLn\) while staying in the same class as \(\curIm\).

\section{Related Work}

\cite{simonyan13_deep_insid_convol_networ} defines a saliency map as the gradient of the network output at the given image.
This would appear to be a sensible definition, but the resulting saliency is very noisy because the network output is roughly constant around any particular image.
\cite{selvaraju16_grad_cam} improves the situation by computing the gradient after each layer instead. This is, however, not a black-box method such as the one we propose.
\cite{koh17_under_black_box_predic_via_influen_funct} computes an \emph{influence function}, that is, a function that measures how the parameters would be changed by a change in the training data. Although it is a black-box method, it is not a saliency method per se.
They use the gradient of the network output to find the pixel most likely to have a high saliency.
The pixels that have most effect are given a higher saliency.

By contrast, \cite{petsiuk18_rise} proposes to directly evaluate the saliency by finding out which pixels are most likely to affect the output, similarly to \cite{fong17_inter_explan_black_boxes_by_meanin_pertur}, but through statistical means instead of iterative optimisation.
These methods can be seen as different ways of solving similar optimisation problems, the solution of which produces a mask (cf. \autoref{sec:imgsAndMasks}) highlighting features of importance.

There are also a number of meta-studies of saliency methods.
\cite{NIPS2018_8160} lists essential properties, for instance the requirement that the results should depend on the training data in a sense that perturbing \emph{model parameters} should change the saliency.
\cite{kindermans17_un_salien_method} proposes a number of properties that saliency methods should satisfy.
\cite{ancona17_towar_better_under_gradien_based} compares several saliency methods and proposes a method to evaluate them (the sensitivity-\(n\) property).

These properties were not in the main focus of the design of our method, but we do fulfill the general criteria. For example, \cite{kindermans17_un_salien_method} is concerned with constant shifts in the inputs. If such a shift is applied to both the target and baseline, then the modification commutes through the interpolation and if we then assume a modified network that deals with such inputs in the same way as the original did with unshifted ones then all the gradients will be the same, therefore the optimised paths will also be the same.

Finally, \cite{Chockler2021occlusion} developed a method to recursively mask out ever finer ``superpixels''. Their construction claims to be based on solid causality principles, and it does seem to largely avoid the adversarial issues that pertubation- / gradient- / statistics-based methods all have to content with (including ours), however the actual implementation approximates the principle in a nondeterministic, mathematically quite unclear way, and the technical details are highly pertaining to the image-classification application. (Still it is a black-box saliency method.)

\section{Ablation Paths}
\label{sec:ablationpath}


\subsection{Images and Masks}
\label{sec:imgsAndMasks}

We assume data is represented by functions on a compact domain \(\Omega\).
Examples for \(\Omega\) may be \(\Omega = \set{1,\ldots,n} \times \set{1,\ldots,m}\) (for pixel images), or a continuous domain \(\Omega = [0,a] \times [0,b]\).
What matters to us is that \(\Omega\) is equipped with a positive measure.
Without loss of generality, we assume the mass of that measure is one, i.e., \(\int_\Omega 1=1\).

The data itself consists of functions on \(\Omega\) with values in a vector space \(\Vect\) (typically, the dimensions of \(\Vect\) may be the \emph{colour channels}).
For the space of images, we choose 
\[
  \Mod \coloneqq \mathcal{C}(\Omega, \Vect)
  .
\]

For our method to work, we need a space of \emph{masks}, denoted \(\Rng\), whose role is to select features between \(\curIm\) and \(\basLn\).
We associate to each mask \(\theta\in\Rng\) an \emph{interpolation operator} between two images \(\curIm\) and \(\basLn\), denoted by \(\Interp{\theta}\).
This interpolator should have the property that \(\Interp{0} = \curIm\) and \(\Interp{1} = \basLn\).
We will thus use the shorthand notation:
\[
  \Interp*{\theta} := \Interp{\theta}
  \qquad
  \theta \in \Rng
  \quad
  (\curIm,\basLn \in \Mod)
  .
\]

The specific choice of masks and interpolation we make in this paper is
\[
  \Rng \coloneqq \Cont(\Omega, \RR)
  ,
  \qquad
  \Interp{\theta} \coloneqq (1-\theta) \curIm + \theta \basLn
  ,
  \qquad
  \theta \in \Rng,
  \quad
  \curIm,\basLn \in \Mod
\]


Another example of interpolation is what in \cite{fong19_understand_dnn_extremperturb_smoothmasks} is called the \emph{pyramid} of \emph{blur} pertubations.
We also explored this possibility for our method, but
we obtain better results with the affine interpolation%
\footnote{Called ``fade pertubation'' in \cite{fong19_understand_dnn_extremperturb_smoothmasks}.}.
(A possible reason that this blur perturbation technique did not work well is the start state, which in that case contains none of the highest-frequency components at all. This makes it likely for the classifier to behave completely different from the target input.)

\subsection{Ablation Paths}\label{monotonePaths}

\begin{definition}
\label{defablationpath}
We define the set
\(
  \Abl
\)
of \emph{ablation paths} as the set of functions \(\pth\colon [0,1] \to \Rng\) fulfilling the following properties:
  \begin{description}
  \item[Boundary conditions]
    \(\pth(0) = 0\) and \(\pth(1) = 1\) 
  \item[Monotonicity]
    \(t_1 \leq t_2 \implies \pth(t_1) \leq \pth(t_2) \qquad t_1,t_2\in[0,1]\)
  \item[Constant speed]
    \(\int_{\Omega} \pth(t) = t \qquad t \in [0,1]\). 
  \end{description}

We will call \emph{monotone paths} the paths that fulfill the first two conditions but not the third.
\end{definition}

Note that the set \(\Abl\) of ablation paths is a \emph{convex subset} inside the set of possible paths denoted by \(\Pth := \Lspace{\infty}([0,1], \Rng)\).

Some comments on each of those requirements are in order.
\begin{enumerate*}[label=\textup{(\roman*)}]
\item 
  \(0\) and \(1\) denote here the constant functions zero and one
  (which corresponds to the zero and one of the algebra \(\Rng\))
\item
  \(\pth(t_1) \leq \pth(t_2)\) should be interpreted as usual as \(\pth(t_2) -\pth(t_1)\) being pointwise nonnegative.
 \item
   If \(t \mapsto \int_{\Omega}\pth(t)\) is differentiable, this requirement
   can be rewritten as \(\frac{\dd}{\dd t}\int_{\Omega} \pth(t) = 1\),
   so it can be regarded as a \emph{constant speed} requirement.
   This requirement is more a normalisation convention than a necessity, as is further detailed in \autoref{rkderpath}.
\end{enumerate*}

The simplest (requirement-fulfilling) ablation path  is the \emph{affine interpolation path}:
\begin{equation}
  \label{eq:interpath}
  \linpath(t) \coloneqq t
  .
\end{equation}
The mask is thus constant in space at each time \(t\). This path is implicitly used in \cite{DBLP:journals/corr/SundararajanTY17}: its image-application corresponds to affine interpolation between target- and baseline image.

\newcommand*\Masses{\mathsf{M}}

  Note that an ablation path without the constant-speed property can always be transformed into one that fulfils it.
  The proof is in \autoref{sec:constantspeed}.

  \begin{remark}
    In the sequel, we will abuse the notations and write \(\pth\) as a function of one or two arguments depending on the context,
    that is, we will identify
    \(
    \pth(t) \equiv \pth(t,\cdot)
    \).
    For instance, in the definition \autoref{defablationpath} above, 
    \(
    \int_{\Omega} \pth(t) \equiv \int_{\Omega} \pth(t,\cdot) \equiv \int_{\Omega}\pth(t,\vcr)\:\dd\vcr.
    \)
  \end{remark}

\begin{remark}
  \label{rkderpath}
  If the ablation path \(\pth\) is differentiable in time, the requirements in \autoref{defablationpath} admit a remarkable reformulation.
  Define \(\psi(t) \coloneqq \frac{\dd}{\dd t}{\pth}(t)\).
  All the requirements in \autoref{defablationpath} are equivalent to the following requirements for a function \(\psi \colon [0,1]\times\Omega \to \RR\):
    \[\psi(t,\vcr) \geq 0 
      ,
\quad
    \int_{\Omega} \psi(t,\vcr) \:\dd\vcr = 1 
    ,
    \quad
    \int_{[0,1]} \psi(t, \vcr) \:\dd t = 1 
    \qquad t\in[0,1],\vcr\in\Omega
    \]
  The corresponding ablation path \(\pth\) is then recovered by \(\pth(t) \coloneqq \int_0^t \psi(s) \,\dd s\).
\end{remark}

\section{Score of an Ablation Path}
\label{sec:score}


We now define the \emph{retaining score function} \(\ScoreRet \colon \Pth \to \RR\) from paths to real numbers by the integral
\begin{equation}
    \label{eq:defscoreRet}
  \ScoreRet(\pth) \coloneqq \int_0^1 F(\Interp*{\pth(t)}) \dd t.
\end{equation}
Note that, as \(F\) is bounded between zero and one, so is \(\ScoreRet(\pth)\) for any ablation path \(\pth \in \Abl\).
The main idea here is that \(F(\curIm) \simeq 1\) and \(F(\basLn)\simeq 0\), and \(F(x)\leq1\) holds always.
So a high value of \(\ScoreRet\) means that the classification stays similar to that of \(\curIm\) over most of the ablation path, which is another way of saying that the characteristics of the original image are retained as best as possible whilst other features of the image are ablated away.
See \autoref{sec:stability} for caveats.

Another score to consider is the \emph{dissipating score}
\begin{equation}
    \label{eq:defscoreDiss}
  \ScoreDiss(\pth) \coloneqq 1 - \int_0^1 F(\Interp*{\pth(t)}) \dd t
\end{equation}
which instead takes high values for paths that ablate crucial features for the current classification as quickly as possible.
Optimisation of \(\ScoreDiss\) corresponds roughly to what \cite{fong17_inter_explan_black_boxes_by_meanin_pertur} call ``deletion game'', whereas \(\ScoreRet\) corresponds to their ``preservation game'', the difference to this work being that they optimise individual masks rather than constrained paths.

Intuitively, the first features to be deleted in a \(\ScoreDiss\)-optimal path \(\pthDiss\) should correspond roughly to the ones longest preserved in a \(\ScoreRet\)-optimal path \(\pthRet\), meaning that a feature that is potent at retaining the classification should be removed early on if the objective is to change the classification.
More generally, one would expect \(\pthDiss(t)\) to be similar to \(1 - \pthRet(1-t)\).

We observe this to be often \emph{not} the case: specifically, there are many examples where either the classification is so predominant that it is almost indeterminate what features should be preserved longest (because any of them will be sufficient to retain the classification), or vice versa the classification is so brittle that it is indeterminate which ones should be removed first.
It is however possible to \emph{enforce} features to be considered simultaneously in a sense of their potency to preserve the classification when they are kept, and changing it when removed. This is achieved by optimising a path with the combined objective of retaining for the path itself and dissipating for its opposite: this is expressed by optimising the \emph{contrastive score}
\begin{equation}
    \label{eq:defscoreCtr}
  \ScoreCtr(\pth) \coloneqq \ScoreRet(\pth) + \ScoreDiss(1 - \pth).
\end{equation}
This too corresponds to ideas already used in previous work, called ``hybrid game'' or ``symmetric preservation'' \cite{fong17_inter_explan_black_boxes_by_meanin_pertur}\cite{Dabkowski2017rtSaliency}.

A related possibility is to train both a retaining and a dissipating path in tandem, but with additional constraints to keep them in correspondence. Here, it is most useful to keep them not opposites of each other, but rather to keep them as similar as possible. 
(Cf. \autoref{fig:decRegSketch}.)
This is achieved by a score of the form
\begin{equation}
    \label{eq:defscoreBndStr}
  \ScoreBndStr(\pthRet, \pthDiss)
      \coloneqq \ScoreRet(\pthRet) + \ScoreDiss(\pthDiss)
      + \lambda_{\pm}
      \norm{\pthRet - \pthDiss}
      ,
\end{equation}
where \(\|\cdot\|\) could refer to various distance notions on the space of paths.
We call the corresponding optimisation problem the \emph{boundary-straddling method}, since (in the ideal of a classifier with exact decision boundaries) it rewards \(\pthRet\) staying in the domain of \(\curIm\) as much as possible and \(\pthDiss\) in the domain of \(\basLn\) as much as possible, i.e. on the other side of the decision boundary but as close as possible.
Thus, \(\pthRet\) and \(\pthDiss\) effectively pinch the decision boundary between them.

For all the above score functions it is straightforward to compute the differential, e.g. \(\dd \ScoreRet\), on the space of paths \(\Pth\):
\[
  \pairing{\dd \ScoreRet}{\delta \pth} = \int_0^1 \pairing{\underbrace{\dd F_{\Interp*{\pth(t)}}}_{\in\Mod^*}}{\underbrace{(\basLn - \curIm)}_{\in\Mod}\underbrace{\delta \pth (t)}_{\in\Rng}} \,\dd t
  \qquad
  \delta\pth \in \Pth
  .
\]

So if we define the product of \(D\in \Mod^*\) and \(x \in \Mod\) producing an element in \(\Rng^*\) by \(\pairing{x D}{\pth} \coloneqq \pairing{D}{x \pth}\) as is customary\footnote{For instance in the theory of distributions.}, we can rewrite this differential as
\[
  \pairing{\dd \ScoreRet}{\delta \pth} = \int_0^1 \pairing[\big]{(\basLn-\curIm)\dd F_{\Interp*{\pth(t)}}}{\delta \pth (t)} \,\dd t
  .
\]
Note that we know that any ablation path is bounded, so \(\pth \in \Lspace{\infty}([0,1], \Rng)\),
so the differential of \(\ScoreRet\) at \(\pth\) can be identified with the function \(\dd\ScoreRet_{\pth} = \bracket[\big]{t\mapsto(\basLn-\curIm) \dd F_{\Interp*{\pth(t)}}}\)
in \(\Lspace1([0,1], \Rng^*)\).

\subsection{Relation with the Integrated Gradient Method}
\label{sec:integratedgrad}

When this differential is computed on the interpolation path \(\linpath\) \eqref{eq:interpath} and then \emph{averaged}, then this is exactly the integrated average gradient \cite{DBLP:journals/corr/SundararajanTY17}.
More precisely, the integrated gradient is exactly \(\int_{0}^1 \dd \ScoreRet_{\linpath(t)} \dd t\).
Note that this is in fact an integrated \emph{differential}, since we obtain an element in the dual space \(\Mod^*\), and this differential should be appropriately smoothed along the lines of \autoref{sec:gradmetric}.


\subsection{Relation to Pixel Ablation}
\label{sec:pixablationTheory}

Given any saliency function \(\sigma \in \Rng\) (for example an integrated gradient, meaningful-pertubation, or grad-CAM result) we can define a path by
\begin{equation}\label{eq:rankingAblation}
  \tilde{\pth}(t) \coloneqq \one_{\sigma \leq \log(t/(1-t))} \text{ when }t\in(0,1)
\end{equation}
and \(\tilde{\pth}(0) \coloneqq 0\), \(\tilde{\pth}(1) \coloneqq 1\).
This path is a {monotone path}, except in the module of images \( \Mod = \Lspace2(\Omega, V)\), equipped with the ring of masks  \(\Rng = \Lspace{\infty}(\Omega)\).
To be an ablation path, it still needs to be transformed into a constant speed path, which is always possible as explained in \autoref{sec:constantspeed}.

That results in a generalisation of the pixel-ablation scores used in \cite{petsiuk18_rise} and \cite{sturmfels2020visualizing}.
In that case, the set \(\Omega\) would be a discrete set of pixels, which are being sequentially switched from ``on'' to ``off'' by the (binary) mask.

Note that in the ranking, pixels with the same saliency would be ranked in an arbitrary way and added to the mask in that arbitrary order.
In the method of \autoref{eq:rankingAblation}, we add such pixels all at once, which seems preferrable because it does not incur an arbitrary bias between pixels. The time reparameterisation keeps the function constant longer to account for however \emph{many} pixels were ranked the same.
As long as the ranking is strict (no two pixels have the same saliency), the method is the same as discrete pixel ranking.

\subsection{Relation to Meaningful Perturbations}
\label{sec:meaningfulpert}

In the saturated case, that is, if \(F\) only takes values zero and one (or in the limit towards this), our method reduces to finding the interpolation with the largest mask on the boundary, equivalent to the approach of \cite{fong17_inter_explan_black_boxes_by_meanin_pertur}.
This is a result of the following: suppose that the ablation path \(\pth\) crosses the boundary at time \(t^*\).
It means that \(F\paren[\big]{\Interp*{\pth(t)}}\) has value one until \(t^*\) and zero afterwards, so the score \(\ScoreRet\)  defined in \eqref{eq:defscoreRet} is \(\ScoreRet(\pth) = t^*\).
By the constant speed property, \(t^* = \int_{\Omega}\pth(t^*)\), so we end up maximising the mask area on the boundary.

\subsection{Relation with RISE}
The method used in \cite{petsiuk18_rise} does not explicitly involve an optimisation problem like here, though they do use pixel ablation as some validation for the results.
It does nevertheless resemble specifically the boundary-straddling method in the sense that it evaluates \(F\) for many different inputs on both sides of the decision boundary, and uses the classification results to weigh the features involved.

\section{Optimisation Problem and Algorithm}\label{sec:optimisation-idea}

We proceed to define the optimisation problem that we propose as a saliency method, and how to solve it numerically.

Conceptually we try to find the ablation path[s] (see \autoref{defablationpath}) that maximises one of the scores \(\ScoreRet(\pth)\), \(\ScoreDiss(\pth)\), \(\ScoreCtr(\pth)\), or \(\ScoreBndStr(\pthRet, \pthDiss)\):
\[
  \max_{\pth\in\Abl} P(\pth)
  .
\]
Recall that the set \(\Abl\) of ablation paths is convex; however, since any of objective functions are not convex, this is not a convex optimisation problem.

The method we suggest is to follow a gradient direction. Such an approach is in general not guaranteed to approximate a global maximum, a common problem with many practical applications. However, empirical results (see \autoref{sec:pointinggameintro}) suggests that gradient descent does often manage to approximate global maxima, particularly obvious in the unregularised near-perfect scores.

\subsection{Gradient and Metric}
\label{sec:gradmetric}

In order to perform gradient descent, we need to be able to compute gradient vectors (sometimes called ``natural gradients'' in the literature).
Strictly speaking the \emph{differential} is an element of the dual space \(\Rng^*\).
In the Euclidean case that space is canonically isomorphic to \(\Rng\), thus the common practice to use it directly as a \emph{gradient} in \(\Rng\), which is usable as contribution to a state update.
In general, this requires first a map from that space to \(\Rng\), and even in a discretised realisation it is prudent to consider this map explicitly, since the implied one depends on the (pixel) basis choice.
A reasonable choice is the covariance operator associated to a smoothing operation.
For a measure \(\mu \in \Rng^*\),
\(
  \pairing{K \mu}{\theta} \coloneqq \pairing{\mu}{\int_{\Omega}k(\cdot - \vcr)\theta(\vcr) \,\dd \vcr}
\),
where \(k\) is a suitable smoothing function. We use here the same Gaussian blurring filter that is also applied between the optimisation steps for regularisation.

Since the optimisation problem is \emph{constrained} (the ablation path \(\pth\) being constrained by the requirements in \autoref{defablationpath}), following the gradient direction will in general leave the set \(\Abl\).
Because the constraints are convex, it is straightforward enough to project each gradient-updated version back to something that does fulfill them, and indeed that is the idea behind our algorithm (see \autoref{sec:pathopt-algo-details} for the technical details).
However, in addition to the hard constraints there are also properties that are desirable but cannot directly be enforced. This is the subject of the next section.

\subsection{Mask saturation}
\label{sec:maskSaturation}
Recall that the masks we use in this paper are functions \(\theta\colon \Omega \to [0,1]\).
The interpretation is that if \(\theta(\vcr) = 0\), the pixel \(\vcr\in\Omega\) of the reference image \(\curIm\) is used, whereas if \(\theta(\vcr)=1\),
the pixel \(\vcr\) of the baseline image \(\basLn\) is used instead.
Typically though, masks take value between zero and one.
%
We notice that such intermediate values of masks produce blending of different images which lie far away from the natural distribution of images.
What is problematic is that the classifier may put such blends of images in totally different classes.
{As analogy in human vision, an image of a person half-blended into an image of a hallway would not be seen as person present at \SI{50}{\percent}, but rather as something completely different; for instance, to a human, this half-present person would look more like a {ghost} than a person.}

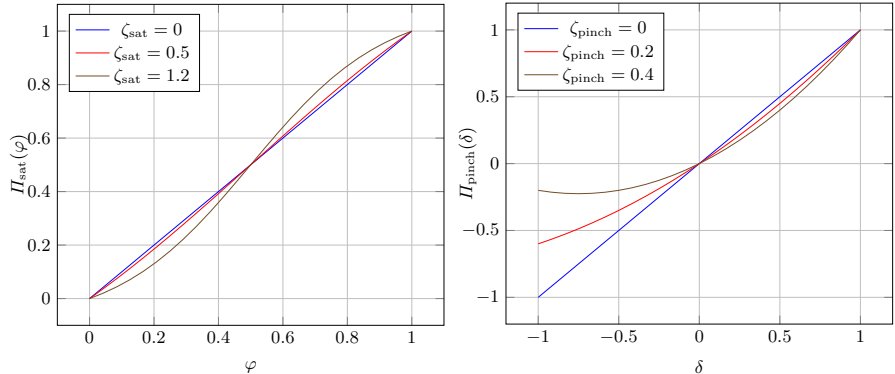
\begin{figure}
  \centering
 \begin{tikzpicture}[domain=0:1, scale=0.75]
    \begin{axis}[
        yticklabel style={%
          scaled ticks=false,
        }
      , grid=major
      , no markers
      , legend pos=north west
      , xlabel=\(\pth\)
      , ylabel=\(\Pi_\text{sat}(\pth)\)
      , y label style={at={(axis description cs:0.12,.5)},anchor=south}
      ]
      \addplot{x};
      \addlegendentry{\(\satStrength=0\)};
      \addplot{(tanh((x*2 - 1)*0.5) / tanh(0.5) + 1)/2};
      \addlegendentry{\(\satStrength=0.5\)};
      \addplot{(tanh((x*2 - 1)*1.2) / tanh(1.2) + 1)/2};
      \addlegendentry{\(\satStrength=1.2\)};
    \end{axis}
 \end{tikzpicture}
 \begin{tikzpicture}[domain=-1:1, scale=0.75]
    \begin{axis}[
        yticklabel style={%
          scaled ticks=false,
        }
      , grid=major
      , no markers
      , legend pos=north west
      , xlabel=\(\delta\)
      , ylabel=\(\Pi_\text{pinch}(\delta)\)
      , y label style={at={(axis description cs:0.12,.5)},anchor=south}
      ]
      \addplot{x};
      \addlegendentry{\(\pinchStrength=0\)};
      \addplot{0.8*x + 0.2*x^2};
      \addlegendentry{\(\pinchStrength=0.2\)};
      \addplot{0.6*x + 0.4*x^2};
      \addlegendentry{\(\pinchStrength=0.4\)};
    \end{axis}
 \end{tikzpicture}
  \caption{The pointwise soft-projection functions for saturation and pinching.}
  \label{fig:softProjections}
\end{figure}

In order to alleviate this potential problem, our algorithm intersperses gradient descent in \(\Pth\) with both (hard) projections onto \(\Abl\), as well as \emph{soft projections} of the masks onto \(\BoRng\),
the set of \emph{saturated masks}, that is, masks taking value in the set \(\set{0,1}\).
Concretely, this is done by tweaking the ablation path pointwise with a sigmoidal function\footnote%
{The exact definitions of \(\Pi_\text{sat}\) and \(\Pi_\text{pinch}\) are largely arbitrary, what matters are their attractive fixpoints; see \autoref{fig:softProjections}}
that brings values lower than \(\tfrac12\) slightly closer to \(0\), and values greater than \(\tfrac12\) slightly closer to \(1\).
\begin{equation}
  \pth \leftarrow \Pi_\text{sat}(\pth) := \frac12
           \left(\frac{\tanh((\pth\cdot2 - 1)\cdot\satStrength)}{\tanh(\satStrength)} + 1\right).
\end{equation}
The parameter \(\satStrength\) determines how strongly this affects the path.

\subsection{Boundary-pinching}
\label{sec:boundaryPinching}
For the boundary-straddling method there is another requirement: making \(\pthRet\) and \(\pthDiss\) similar to each other can be achieved by explicitly penalizing their distance in the score function, but in our implementation this too is done by a dedicated algorithm step that manipulates the masks pointwise to become more similar.
For interpretability purposes it is particularly desirable for \(\pthRet(t)\) to contain only few features that \(\pthDiss(t)\) does not, since that allows direct comparison between two images showing how inclusion of these features bring the classification into the target class.
The exact difference in strength of features meanwhile is less relevant (even when the masks themselves are not boolean). Accordingly, we suggest a \emph{pinching tweak} that diminishes specifically the smaller positive differences between \(\pthRet\) and \(\pthDiss\), in addition to any negative differences. The concrete form in our experiments is this: (recall that values close to 1 correspond to masked-\emph{away} features)
\begin{equation}
 \label{eq:asymPinch}
 \pthDiss(t,\vcr) \leftarrow \pthRet(t,\vcr)
                   + \Pi_\text{pinch}\bigl(\pthDiss(t,\vcr) - \pthRet(t,\vcr)\bigr)
\end{equation}
where \(\Pi_\text{pinch}: [-1,1]\to[-1,1], \delta\mapsto\Pi_\text{pinch}(\delta)\) is a continuous function with an attractive fixpoint at \(\delta=0\) (which is responsible for squelching unsubstantial contrasts between \(\pthRet\) and \(\pthDiss\)), and a repulsive one at \(\delta=1\) (which allows the most salient features of \(\pthRet\) to stay absent from \(\pthDiss\), as necessary for a high \(\ScoreBndStr\)).

The concrete definition of \(\Pi_{\text{pinch}}\) is uncritical, in our experiments we used
\[
      \Pi_{\text{pinch}}(\delta)
        := \delta(1 - \pinchStrength) + \delta^2 \pinchStrength.
\]
Notice that in \autoref{eq:asymPinch}, \(\pthRet\) is not affected by \(\pthDiss\), only vice versa. But conceptually, the update is performing a change to \(\delta\), i.e. the difference between the paths, rather than either of them individually.

\section{Stability and adversarial effects}
\label{sec:stability}

So far it was more or less taken for granted that a high-scoring ablation path owes its score to good highlighting of the features that were also responsible for the classification of \(\curIm\).
This assumption would be reasonable if any masked version of that image were classified either the same for the same reasons as the original, or else classified differently. For a black-box model, there is however no way of verifying this, and in fact it is simply not true in general.

It is well known \cite{fong17_inter_explan_black_boxes_by_meanin_pertur} that sufficiently pathological masks can act as \emph{adversarial attacks} \cite{szegedy2014intriguing} on an image, i.e. that masking out very minor parts of an image may affect the classification disproportionally and in ways that involve completely different neural activations (or whatever other concept is appropriate for the classifier architecture at hand).
Gradient descent approaches tend to produce such examples easily.
Although the study of adversarial effects is an important matter of its own right, they are hardly relevant for saliency purposes, because they do not necessarily involve the features that caused the classification of the original image.

A standard technique \cite{fong17_inter_explan_black_boxes_by_meanin_pertur}\cite{petsiuk18_rise} employed to avoid that masks affect images adversarially is to regularize them in some sense of smoothness, e.g. by adding a total variation penalty.
Intuitively, this at least prevents the masked image from featuring sharp edges or similar details not present in \(\curIm\) that the classifier might latch onto.
We implemented this in terms of a simple Gaussian filter applied to each mask in the path after each optimisation step.
Empirically, strong enough smoothing does largely avoid adversarial classification, but unfortunately there is no a-priori way of telling how smooth it needs to be. And too strong smoothing can also have detrimental effects. Not only does it prevent the exact localization of small, salient features, but it can even bias the outcome:
\begin{figure}
 \centering
 \begin{tabular}{ccc}
  \includegraphics[width=0.3\textwidth]{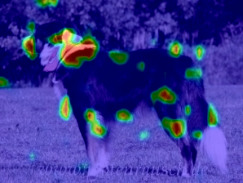}
  &
  \includegraphics[width=0.3\textwidth]{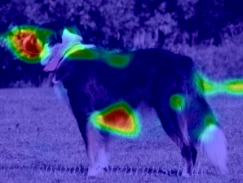}
  &
  \includegraphics[width=0.3\textwidth]{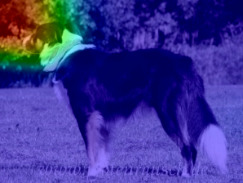}
   \\
   \(\sigma_\text{reguBlur} = 0.5\)
    &
   \(\sigma_\text{reguBlur} = 2\)
    &
   \(\sigma_\text{reguBlur} = 8\)
 \end{tabular}
 \caption{Example of how both too little and too much regularisation can be detrimental for interpretability. Image from VOC2007 test set; saliency is class transition of a \(\ScoreCtr\)-optimal path.}
 \label{fig:reguDilemma}
\end{figure}
in \autoref{fig:reguDilemma}, the strongly regularised saliency is not only condensed to a single location, but also specifically to a corner of the image. Out interpretation is that this happens because it reduces the total variation (since \(\tfrac34\) of the gradient of the mask lies outside of the image). And although the mask still contains enough of the dog's head to keep the classification, its maximum lies misleadingly in front of the nose.

Adversarialness, or generally instability of the mask-classification interaction, can also be viewed in terms of a decision boundary that is fractal-like crinkled in the high-dimensional image space, such that small exclaves of a class domain may reach far closer to \(\curIm\) than the bulk of that domain. This suggests that it would help to evaluate for many different masks rather than gradient-optimising a single one.
Particularly the RISE method \cite{petsiuk18_rise} benefits from this, by evaluating the classifier for a whole large random selection of masks. The reference implementation of \cite{fong19_understand_dnn_extremperturb_smoothmasks} also optimises (individually) multiple masks of different sizes. This does not so much avoid adversarial examples as average out their contributions, whereas stable, faithfully-salient masks tend to agree.
(This still requires also dedicated mask regularity, as otherwise the adversarial contributions overwhelm and the result appears as mere noise.)


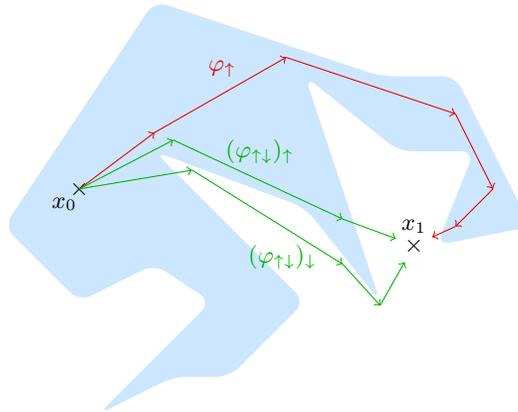
\begin{figure}[htb]
 \centering
 \begin{tikzpicture}
    \definecolor{curTgtDomCol}{rgb}{0.8,0.9,1}

    \node [coordinate] (0) at (-4, 1) {};
    \node [coordinate] (1) at (-2, 0) {};
    \node [coordinate] (2) at (-1, 1) {};
    \node [coordinate] (3) at (-3, 3) {};
    \node [coordinate] (4) at (0, 1) {};
    \node [coordinate] (5) at (-1, 2.25) {};
    \node [coordinate] (6) at (2, 2) {};
    \node [coordinate] (7) at (1, 4) {};
    \node [coordinate] (8) at (-1, 4) {};
    \node [coordinate] (9) at (0, 4) {};
    \node [coordinate] (10) at (-3, 5) {};
    \node [coordinate] (11) at (-5, 2) {};
    \node [coordinate] (12) at (0.75, 1.75) {};
    \node [coordinate] (13) at (1.25, 3) {};
    \node [coordinate] (14) at (0.25, 3) {};
    \node [coordinate] (15) at (-2.75, -0) {};
    \node [coordinate] (16) at (-2.25, 1) {};
    \node [coordinate] (17) at (-3.75, -0.5) {};
    \node [coordinate] (18) at (-3, 1.5) {};

    \path [rounded corners=5pt, fill=curTgtDomCol]
          (0) -- (18) -- (16) -- (17) -- (15)
              -- (1) -- (2) -- (3) -- (5) -- (4)
              -- (8) -- (14) -- (13) -- (12) -- (6)
              -- (7) -- (9) -- (10) -- (11) -- cycle;

    \node (X) at (-4, 2.5) {$\times$};
    \node (Xlbl) at (-4.2, 2.3) {$\curIm$};
    \node (BL) at (0.45, 1.75) {$\times$};
    \node (BLlbl) at (0.45, 2) {$\basLn$};

    \definecolor{retPathCol}{rgb}{0.9,0,0}
    \definecolor{stradPathsCol}{rgb}{0,0.7,0}

    \node [coordinate] (21) at (-3, 3.25) {};
    \node [text=retPathCol] (RPlbl) at (-2.1, 4.1) {$\pthRet$};
    \node [coordinate] (22) at (-1.25, 4.25) {};
    \node [coordinate] (23) at (1, 3.5) {};
    \node [coordinate] (24) at (1.5, 2.5) {};
    \node [coordinate] (25) at (1, 2) {};
    \node [coordinate] (26) at (-2.75, 3.15) {};
    \node [text=stradPathsCol] (SPPlbl) at (-1.6, 3) {$(\pthBndStr)_\uparrow$};
    \node [coordinate] (27) at (-0.5, 2.1) {};
    \node [coordinate] (28) at (-2.5, 2.75) {};
    \node [text=stradPathsCol] (SPMlbl) at (-1.3, 1.6) {$(\pthBndStr)_\downarrow$};
    \node [coordinate] (29) at (-0.5, 1.5) {};
    \node [coordinate] (30) at (0, 0.95) {};

    \draw [->, retPathCol] (X.center) -- (21);
    \draw [->, retPathCol] (21) -- (22);
    \draw [->, retPathCol] (22) -- (23);
    \draw [->, retPathCol] (23) -- (24);
    \draw [->, retPathCol] (24) -- (25);
    \draw [->, retPathCol] (25) -- (BL);

    \draw [->, stradPathsCol] (X.center) -- (26);
    \draw [->, stradPathsCol] (26) -- (27);
    \draw [->, stradPathsCol] (27) -- (BL);

    \draw [->, stradPathsCol] (X.center) -- (28);
    \draw [->, stradPathsCol] (28) -- (29);
    \draw [->, stradPathsCol] (29) -- (30);
    \draw [->, stradPathsCol] (30) -- (BL);
\end{tikzpicture}
 \caption{Low-dimensional sketch of how irregularities in the decision boundary between two images can affect optimised paths. The \(\ScoreRet\)-optimised one takes a detour along a strong outlier ($\approx$adversarial) that allows it to stay almost completely in-domain, which causes the final section to come from a completely different direction in the end.
   The pair of \(\ScoreBndStr\)-optimised paths are still somewhat affected by outliers, but the pinching term causes them to mostly follow a more regular and consistent section of the boundary.\\
   N.B.: this represents only very crudely the behaviour in real image classification applications, as inevitable with low-dimensional visualisations. In particular, the monotonicity condition is not represented at all here, and the pinching is here a symmetric \(L^2\)-reduction, which is quite different from \autoref{eq:asymPinch}.}
 \label{fig:decRegSketch}
\end{figure}

It appears that the irregularity of the decision boundary is better described as thin outreaching folds rather than standalone islands (\autoref{fig:decRegSketch}), such that even a monotone path can follow them up to an adversarial point before crossing the decision boundary very near \(\curIm\). The scores \(\ScoreRet\) and \(\ScoreDiss\) are particularly prone to procuring such paths, but we observe them also when optimising for \(\ScoreCtr\).

By comparison, \(\ScoreBndStr\) seems to be more reliable in practice. An intuitive reason is that the two paths have less possibility to simultaneously follow adversarial masks for the classes of both \(\curIm\) and \(\basLn\), whilst also staying close to each other. That is certainly not inconceivable either, though.

\section{Pointing game}
\label{sec:pointinggameintro}
We evaluate our saliency algorithm using the pointing game.
This method was introduced in \cite{zhang17_top_down_neural_atten_by_excit_backp} and used, for instance, in \cite{selvaraju16_grad_cam}\cite{fong17_inter_explan_black_boxes_by_meanin_pertur}.
It checks whether the maximum pixel of a saliency heatmap agrees with the location of a  human-annotated%
\footnote{The pointing game is generally used under the assumption that neither the classifier nor the saliency method have any direct training knowledge about the position annotations, i.e. it is not a test of how well a trained task generalizes but of an extrinsic notion of saliency.
} object of the class of interest.

Assessments like the pointing game have their caveats for benchmarking saliency methods.
One can for instance argue that the cases when the saliency points somewhere outside the bounding box are the most insightful ones, as they indicate that the classifier is using information from an unexpected part of the image (for instance, the background).
Another caveat is that, if winning at the pointing game is the goal, a saliency method is only as good as its underlying classifier is.
For these reasons, a pointing game score should not be considered as the predominant criterion for a good saliency method.

Nevertheless, it is reasonable to expect a useful saliency method to perform at least similarly well as the state of the art: if existing methods have proven capable of achieving high scores, this is after all indication that the classifier does to a significant degree base its decisions on spatially confined features of the real objects.
Furthermore the pointing game provides a way of comparing the behaviour of different variations of a saliency method (such as different hyperparameters) somewhat more representatively than looking at individual image examples. Again, the best-scoring parameters are not necessarily the best for attribution purposes, but they are a reasonable starting point.

\subsection{Heatmap reduction}
The result of the methods introduced in this paper is one or multiple paths, whereas the pointing game expects a single heatmap. There are multiple ways of reducing to such a map:
\subsubsection{Averaging.}
One can simply average over all the masks in a \(\ScoreRet\)-optimal path. This operation is (modulo a time renormalisation) left inverse to the pixel ablation of a saliency map (\autoref{sec:pixablationTheory}).
\begin{equation}
 \label{eq:naiveAvgMask}
  \overline{\pthRet} := \int_0^1\pth(t)\dd t.
\end{equation}
This works well in some cases, but the result can be disproportionally affected by low-discriminate contrasts of masks generated far from a decision boundary, which are unstable in a similar way to plain gradient methods.

\subsubsection{Class transition.}
\label{sec:classtrans}
Taking the point of view that the decision boundary is what matters, one can seek the position where the path crosses it by tracking the classifier outputs along the path.

Empirically, this gives better results than averaging (both for the pointing game and, to our eyes, ease of interpretation), but it hinges on the assumption of there being a single boundary-crossing. In general, there may be multiple crossings, or the classifier might have a far more gradual transition, or (in case an explanation for a class different from the prediction for \(\curIm\) is sought) it might not cross a boundary at all.
In our implementation, we therefore make a case distinction:
\begin{itemize}
 \item If there exist \(t\) such that \(F(\pth(t))\) is dominated by the target class, then we select the largest of these \(t\). In other words, we select the most confined mask that results in the classification of interest.
  Here (unlike the rest of the paper) we consider the full multi-class output of \(F\), and by ``dominate'' we mean that the target class ranks higher than all others.
 \item If no such \(t\) exists, we select simply \(\argmax_t F(\pth(t))\).
\end{itemize}
This may not be the best strategy in all applications, but it does guarantee always getting a result that can be compared in the pointing game. In critical applications it is likely better to discard paths that do not cross a boundary, and consult a different method in such a case.

\subsubsection{Contrastive averaging.}
\label{sec:contrastiveAvg}
For the two paths optimizing \(\ScoreBndStr\), the property of interest is that they pinch the decision boundary between them. That means that for each \(t\), the normal direction of the boundary is approximated by \(\pthRet(t) - \pthDiss(t)\) (at least coarsely, cf. \autoref{fig:decRegSketch}). This suggests averaging between these values, i.e.
\begin{equation}
 \label{eq:bndStraAvgMask}
  \overline{\pthBndStr} := \int_0^1\bigl(\pthRet(t) - \pthDiss(t)\bigr)\dd t.
\end{equation}
Indeed this appears to give comparatively good, stable results in practice. Our interpretation is that on any indiscriminate parts%
of the path, the pinching tweak \autoref{eq:asymPinch} reduces \(\pthRet(t) - \pthDiss(t)\) so these parts do not contribute to the result like they would in \autoref{eq:naiveAvgMask}.
The reason for this behaviour is that indiscrimate parts do not have a consistent \(F\)-gradient that would keep \(\pthRet(t)\) and \(\pthDiss(t)\) apart during optimisation. On the other hand, stably-salient differences do keep them apart and therefore prevail in \(\overline{\pthBndStr}\).

\subsection{Results}
\label{sec:pointinggameresults}

We initially ran a custom implementation of the pointing game on individual classes (synsets) from the ImageNet dataset. In some cases even simple \(\ScoreRet\)-optimal paths perform well, e.g. \SI{83}{\percent} on the ``Bee'' synset with EfficientNet classifier, which is better than the result with saliency methods from the literature. 
These experiments turned however out to be not very representative: on larger and mixed datasets, we were not able to find hyperparameters that avoided high unstability in the optimisation and consequently lower scores, especially in case of \(\ScoreRet\).

For fair and representative comparison with the literature, we present here the result on a benchmark that was already used in 
\cite{fong19_understand_dnn_extremperturb_smoothmasks}. Specifically, we used their TorchRay suite \cite{TorchRay} to evaluate the saliency-result of our method (reduced to a heatmap), explaining the classifications by ResNet50 on the COCO14 validation dataset. 
We did not have the computational resources to run the whole set, so used the first 1000 images%
\footnote{By ``first 1000'' we mean the 1000 images with the lowest ids. Note that the VOC and COCO sets are in random order, so that this should be a reasonably representative and reproducible selection. Comparing the score of Grad-CAM to the one on the full datasets confirms this.\\
The astute reader may notice that on the other hand, with only the first 100 images the results are systematically worse. This is less due to these images being more difficult, than artifact of the way the TorchRay benchmark gathers the results: specifically, it counts success rate for each class separately and averages in the end, but rates classes that are not even present in the smaller subset as 0\% success.}
for a comparison to the literature state of the art of our best-scoring result (\autoref{tab:literatureCompare}), which was in turn determined among variations of our method on the first 100 images (\autoref{tab:varietiesCompare}).
For each image, a saliency is obtained for each of the annotated objects, so for example the 100 COCO images correspond to 310 optimised paths.

\begin{table}
  \centering
  \begin{tabular}{c|cc}&\textit{VOC07 Test}&\textit{COCO14 Val}\\
Method&(All\%/Diff\%)&(All\%/Diff\%)\\\hline 
Ctr.&70.9/41.9&26.0/15.4\tabularnewline 
GCAM&90.5/80.4&57.1/49.2\tabularnewline 
Ours&84.3/64.8&49.3/41.0\end{tabular}
  \begin{tabular}{||c|cc}&\textit{VOC07 Test}&\textit{COCO14 Val}\\
Method&(All\%/Diff\%)&(All\%/Diff\%)\\\hline 
RISE&86.4/78.8&54.7/50.0\tabularnewline 
GCAM&90.4/82.3&57.3/52.3\tabularnewline 
Extr&88.9/78.7&56.5/51.5\end{tabular}

  \caption{\emph{Left}: the highest-scoring results for the pointing game over 1000 images with ResNet50 classifier, for comparison with the state of the art. \emph{Right}: excerpt from table 1 in \cite{fong19_understand_dnn_extremperturb_smoothmasks}, which contains the scores of more methods from the literature for the complete datasets.}
  \label{tab:literatureCompare}
\end{table}

The top scores are close to the state of the art, but do not quite reach the pointing accuracy of Grad-CAM, nor of extremal pertubation or RISE.
It is not clear whether this is a result of fundamental limitations of our approach, of remaining stability problems that could be fixed with e.g. other regularisation approaches, or whether it is even a deficiency at all.
Clearly the method does work in principle, and it is by construction faithful, so the somewhat higher mismatch rate could also be construed as higher sensitivity to aberrant or unstable behaviour of the classifier.

\begin{table}
  \centering
  \begin{tabular}{cccccc|cc}&&&&&&\textit{VOC07 Test}&\textit{COCO14 Val}\\
Method&opt.crit&intp.spc&${\zeta{}}_{\text{sat}}$&${\sigma{}}_{\text{reguBlur}}$&postproc&(All\%/Diff\%)&(All\%/Diff\%)\\\hline 
Ctr.&&&&&&71.4/36.6&26.5/11.2\tabularnewline 
GCAM&&&&&&90.4/64.2&48.9/35.2\tabularnewline 
Abl.Path&\hyperref[eq:defscoreRet]{$\ScoreRet{}$}&blur-fade&0.8&8.0px&&44.0/38.1&30.2/23.2\tabularnewline 
Abl.Path&\hyperref[eq:defscoreCtr]{$\ScoreCtr{}$}&blur-fade&0.8&2.0px&&73.8/46.9&38.8/26.4\tabularnewline 
Abl.Path&\hyperref[eq:defscoreCtr]{$\ScoreCtr{}$}&blur-fade&0.8&7.0px&&52.4/40.0&32.5/23.1\tabularnewline 
Abl.Path&\hyperref[eq:defscoreCtr]{$\ScoreCtr{}$}&blur-fade&0.8&7.0px&window&76.4/48.9&40.6/26.7\tabularnewline 
Abl.Path&\hyperref[eq:defscoreCtr]{$\ScoreCtr{}$}&blur-fade&0&7.0px&window&80.5/46.1&46.2/31.0\tabularnewline 
Abl.Path&\hyperref[eq:defscoreBndStr]{$\ScoreBndStr{}$}&blur-fade&0.8&7.0px&&72.2/47.5&48.3/34.8\tabularnewline 
Abl.Path&\hyperref[eq:defscoreBndStr]{$\ScoreBndStr{}$}&pyramid&0.8&8.0px&&75.2/47.1&39.5/26.5\end{tabular}
  \caption{Some of our results for the pointing game over 100 images with ResNet50 classifier.}
  \label{tab:varietiesCompare}
\end{table}

Different variations of our methods also perform quite differently. We cannot describe all the observations that could be made from these experiments here, nor is this necessarily useful (many of the trends here likely do not generalize to other data), but a selection that may be noteworthy:
\begin{itemize}
 \item The simple single-path retaining (or dissipating) methods compete badly. See \autoref{sec:stability} for some possible explanations.
 \item The boundary-straddle method performs best on the COCO dataset, and also relatively good on the difficult subset of VOC. We propose that this is typically the best of our methods, though in particular on the simple subset of VOC its results are quite dissappointing.
 \item The contrastive method performs well in particular on the simple subset of VOC, but only with very particular regularisation settings; see \autoref{sec:pgameHyperparams}. With e.g. strong blurring and saturation but no windowing, it may perform worse than even the trivial center method, evidently an artifact of the effect shown in \autoref{fig:reguDilemma}.
\end{itemize}

\subsection{Hyperparameter choice}
\label{sec:pgameHyperparams}

Most of the saliency methods from the literature have some hyperparameters%
\footnote{The authors in \cite{fong19_understand_dnn_extremperturb_smoothmasks} emphasize that their method avoids hyperparameters, yet their examples rely on no fewer than 5 hard-coded number constants.},
as does ours. The ideal choice of these parameters is little discussed.
Unlike when training a machine learning model, there is no objective on which this choice should unambiguously be based, but it appears that several authors have used the pointing game for this purpose.
Apart from the aforementioned caveats, this also has the problem that the required position-annotations are simply not available for most applications.

In our case, there \emph{is} an additional score with a clear meaning available: the ablation path score. And though it is evidently not the case that the hyperparameters leading to the highest ablation score give the best saliency results, we propose that studying only the ablation score can nevertheless provide some guidance for a good choice.
\begin{figure}[htb]
  \centering
  \includegraphics[width=0.8\textwidth]{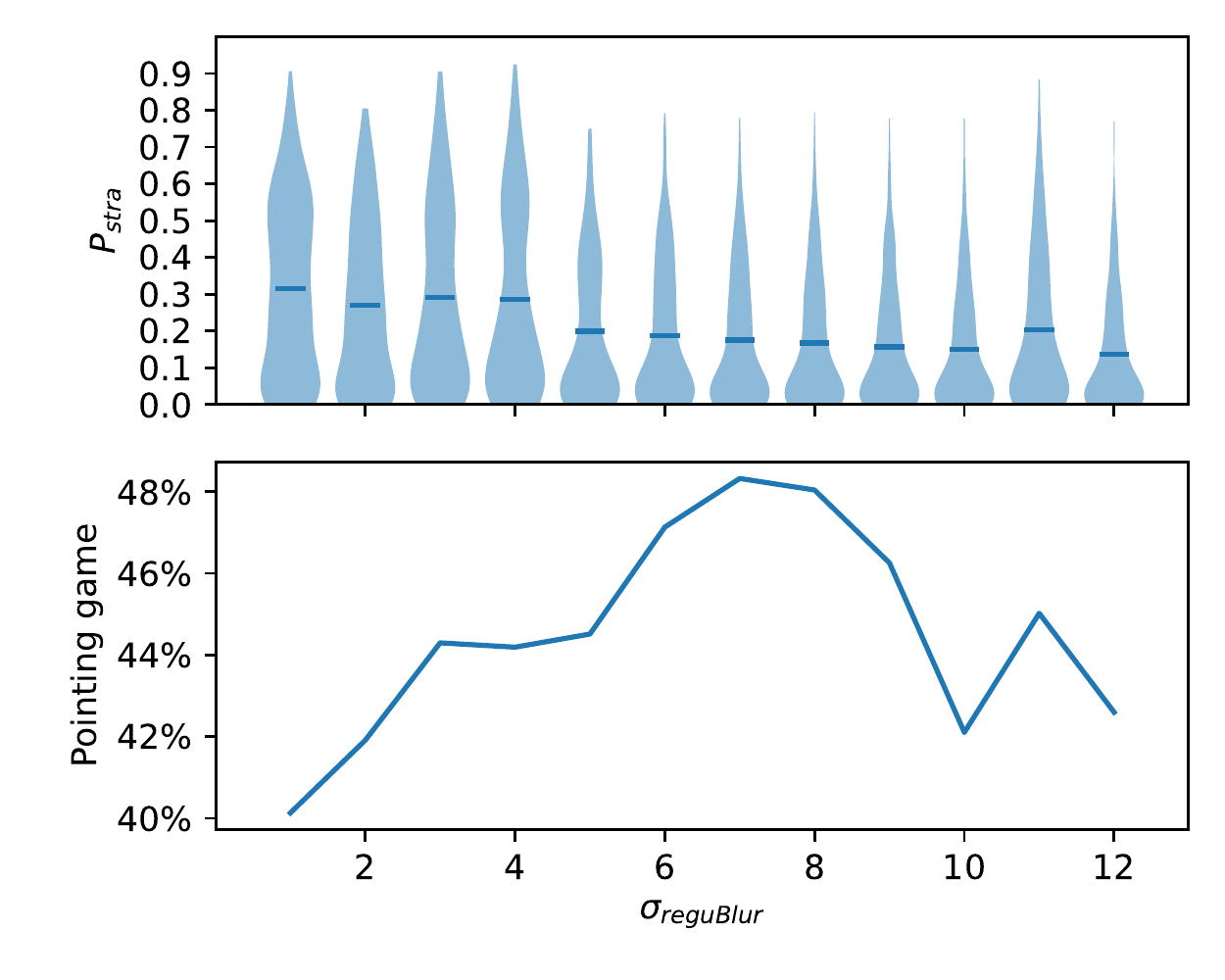}
  \caption{%
   Dependence on the size of the regularization filter, for both the distribution of boundary-straddling ablation-path scores and the pointing game score (evaluated with contrastive averaging as per \autoref{sec:contrastiveAvg}). Based on ResNet50 and 100 images from the COCO14 dataset.%
  }
  \label{fig:pgame-vs-pathscore-strad}
\end{figure}

Specifically, the regularisation parameter is responsible for avoiding adversarial masks, which can be identified by a large population of paths with very high score. Observe in \autoref{fig:pgame-vs-pathscore-strad} that the  histograms at low \(\sigma_\text{reguBlur}\) have an upper bulge (\(\ScoreBndStr>0.5\)). After \(\sigma_\text{reguBlur}\) has been increased to a size of 5 pixels, the adversarial population vanishes, and accordingly the pointing game score rises towards its maximum at \(\sigma_\text{reguBlur}=7\).

Because the path score is a property without application-specific dimension, this phenomen can also be expected in applications where very different regularisation is required compared the image classification examples here.
This is therefore a possible criterion for hyperparameter choice when the pointing game or an analog is not available. Some care needs to be taken though:
\begin{figure}[htb]
  \centering
  \includegraphics[width=0.8\textwidth]{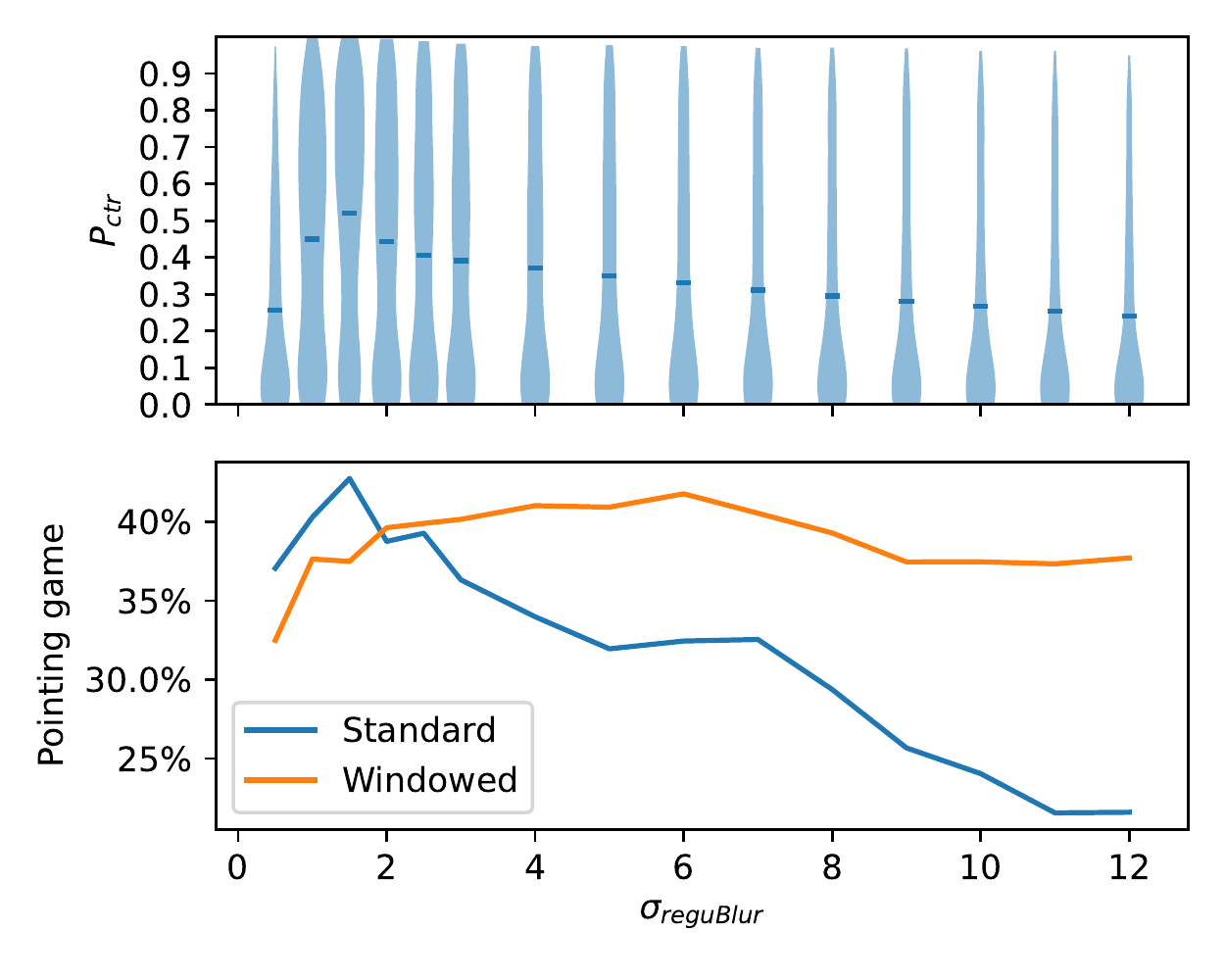}
  \caption{%
    Like \autoref{fig:pgame-vs-pathscore-strad}, but with paths optimised for the contrastive score. Pointing game evaluated on both standard class-transition masks (\autoref{sec:classtrans}), and with a boundary-suppressing window applied.%
  }
  \label{fig:pgame-vs-pathscore-ctr}
\end{figure}
\autoref{fig:pgame-vs-pathscore-ctr} shows an example where the pointing game with unmodified masks has a steadily decreasing score as the regularization is increased. We have identified the problem as stemming from the tendency of the regularization to push the argmax to the image boundary (cf. \autoref{fig:reguDilemma}). This particular effect can be prevented by post-processing the masks with a window that suppresses the boundaries\footnote{%
  This can be interpreted as applying prior knowledge of the location of objects in the dataset. However, there \emph{are} also examples of objects close to the boundary. The window post-processing prevents these from being properly localised, which is why the top pointing-game score is still lower.
}, in which case the rule suggested above is valid again.

\section{Conclusion}

We demonstrated that the ablation path formalism provides a usable saliency method that combines ideas from several previous methods within a single mathematical framework. The ablation path method can stand in for each of these methods to some extent, and is with suitable parameter choices also able to produce results that score similarly well in the pointing game.

This is a nontrivial result, because these methods appear quite different in their original formulations. And even though ours has strong similarity to
\cite{fong17_inter_explan_black_boxes_by_meanin_pertur} / \cite{fong19_understand_dnn_extremperturb_smoothmasks}, it was a priori not obvious that the restriction to a path instead of individual masks would still leave the optimisation problem solvable in practice.
Indeed, for some inputs our method still struggles to converge on a human-reasonable explanation, even when other methods accomplish this. It is possible that in some cases there simply exist no paths that the classifier can follow in a well-behaved way.
But for most of the examples we tested on, this does not seem to be a fundamental issue.

The main practical advantage of an ablation path, which is most evident when interactively browsing through it and tracking the exact classifier response, is the added information: unlike each of the previous methods, an ablation path offers a whole sequence of fine-grained changes to an input image.
It thus offers a more thorough insight to the classification, while still ensuring the explanations form a consistent picture thanks to the monotonicity condition. Because each point in a path is associated with a concrete input to the network whose result can directly be inspected, we argue the method is \emph{faithful}~\cite{weller2019transparency}, whilst also being easy and intuitive to use.
A caveat is that without suitable parameters (in particular regularisation), the explanations may highlight only the adversarial behaviour of a classifier, or even respond to spoofing by a classifier designed to detect the artificial inputs (``Volkswagening''). This possibility is to our knowledge common to all black-box saliency methods, so their use in critical applications should be considered carefully\cite{Rudin_2019}.

A disadvantage of our specific approach, in addition to the stability issues, is the rather high computational effort. A path requires many (ca. 50) optimisation steps, each of which require several classifier evaluations (ca. 20).

The method is perhaps best used in tandem with another one, for example Grad-CAM which is fast and stable but lacks the possibility of assessing faithfulness.

\newpage

\bibliography{bib}

\begin{thebibliography}{10}
\providecommand{\url}[1]{\texttt{#1}}
\providecommand{\urlprefix}{URL }
\providecommand{\doi}[1]{https://doi.org/#1}

\bibitem{NIPS2018_8160}
Adebayo, J., Gilmer, J., Muelly, M., Goodfellow, I., Hardt, M., Kim, B.: Sanity
  checks for saliency maps. In: Bengio, S., Wallach, H., Larochelle, H.,
  Grauman, K., Cesa-Bianchi, N., Garnett, R. (eds.) Advances in Neural
  Information Processing Systems 31, pp. 9505--9515. Curran Associates, Inc.
  (2018),
  \url{http://papers.nips.cc/paper/8160-sanity-checks-for-saliency-maps.pdf}

\bibitem{ancona17_towar_better_under_gradien_based}
Ancona, M., Ceolini, E., {\"O}ztireli, C., Gross, M.: Towards better
  understanding of gradient-based attribution methods for deep neural networks.
  CoRR  (2017)

\bibitem{Chockler2021occlusion}
Chockler, H., Kroening, D., Sun, Y.: Explanations for occluded images. In:
  Proceedings of the IEEE/CVF International Conference on Computer Vision
  (ICCV). pp. 1234--1243 (October 2021)

\bibitem{Dabkowski2017rtSaliency}
Dabkowski, P., Gal, Y.: Real time image saliency for black box classifiers. In:
  Guyon, I., Luxburg, U.V., Bengio, S., Wallach, H., Fergus, R., Vishwanathan,
  S., Garnett, R. (eds.) Advances in Neural Information Processing Systems.
  vol.~30. Curran Associates, Inc. (2017),
  \url{https://proceedings.neurips.cc/paper/2017/file/0060ef47b12160b9198302ebdb144dcf-Paper.pdf}

\bibitem{fong19_understand_dnn_extremperturb_smoothmasks}
Fong, R., Patrick, M., Vedaldi, A.: Understanding deep networks via extremal
  perturbations and smooth masks pp. 2950--2958 (2019)

\bibitem{fong17_inter_explan_black_boxes_by_meanin_pertur}
Fong, R.C., Vedaldi, A.: Interpretable explanations of black boxes by
  meaningful perturbation  (Oct 2017)

\bibitem{kindermans17_un_salien_method}
Kindermans, P.J., Hooker, S., Adebayo, J., Alber, M., Sch{\"u}tt, K.T.,
  D{\"a}hne, S., Erhan, D., Kim, B.: The (un)reliability of saliency methods
  pp. 267--280 (2019). \doi{10.1007/978-3-030-28954-6_14},
  \url{https://doi.org/10.1007/978-3-030-28954-6_14}

\bibitem{koh17_under_black_box_predic_via_influen_funct}
Koh, P.W., Liang, P.: Understanding black-box predictions via influence
  functions  \textbf{70},  1885--1894 (06--11 Aug 2017),
  \url{https://proceedings.mlr.press/v70/koh17a.html}

\bibitem{petsiuk18_rise}
Petsiuk, V., Das, A., Saenko, K.: Rise: Randomized input sampling for
  explanation of black-box models. CoRR  (2018)

\bibitem{Rudin_2019}
Rudin, C.: Stop explaining black box machine learning models for high stakes
  decisions and use interpretable models instead. Nature Machine Intelligence
  \textbf{1}(5),  206--215 (may 2019). \doi{10.1038/s42256-019-0048-x},
  \url{https://doi.org/10.1038%2Fs42256-019-0048-x}

\bibitem{selvaraju16_grad_cam}
Selvaraju, R.R., Cogswell, M., Das, A., Vedantam, R., Parikh, D., Batra, D.:
  Grad-cam: Visual explanations from deep networks via gradient-based
  localization  (Oct 2017)

\bibitem{simonyan13_deep_insid_convol_networ}
Simonyan, K., Vedaldi, A., Zisserman, A.: Deep inside convolutional networks:
  Visualising image classification models and saliency maps. CoRR  (2013)

\bibitem{sturmfels2020visualizing}
Sturmfels, P., Lundberg, S., Lee, S.I.: Visualizing the impact of feature
  attribution baselines. Distill  (2020). \doi{10.23915/distill.00022},
  https://distill.pub/2020/attribution-baselines

\bibitem{DBLP:journals/corr/SundararajanTY17}
Sundararajan, M., Taly, A., Yan, Q.: Axiomatic attribution for deep networks
  \textbf{70},  3319--3328 (06--11 Aug 2017),
  \url{https://proceedings.mlr.press/v70/sundararajan17a.html}

\bibitem{szegedy2014intriguing}
Szegedy, C., Zaremba, W., Sutskever, I., Bruna, J., Erhan, D., Goodfellow, I.,
  Fergus, R.: Intriguing properties of neural networks (2014)

\bibitem{TorchRay}
Vedaldi, A.: Understanding deep networks via extremal perturbations and smooth
  masks. \url{https://github.com/facebookresearch/TorchRay} (2019)

\bibitem{weller2019transparency}
Weller, A.: Transparency: Motivations and challenges. In: Samek, W., Montavon,
  G., Vedaldi, A., Hansen, L.K., M{\"u}ller, K.R. (eds.) Explainable AI:
  interpreting, explaining and visualizing deep learning, vol. 11700, chap.~2.
  Springer Nature (2019)

\bibitem{zhang17_top_down_neural_atten_by_excit_backp}
Zhang, J., Bargal, S.A., Lin, Z., Brandt, J., Shen, X., Sclaroff, S.: Top-down
  neural attention by excitation backprop. International Journal of Computer
  Vision  \textbf{126}(10),  1084--1102 (2017). \doi{10.1007/s11263-017-1059-x}

\end{thebibliography}
\bibliographystyle{splncs04}

\appendix

\renewcommand{\thesection}{Appendix \Alph{section}}

\section{Canonical Time Reparametrisation}
\label{sec:constantspeed}

  \begin{proof}[Proof of \autoref{prop:constantspeed}]
  The function \(m \colon [0,1] \to \RR\) defined by \(m(t) \coloneqq \int_{\Omega} \pth(t)\) is increasing and goes from zero to one (since we assume that \(\int_{\Omega} 1 = 1\)).

  Note first that if \(m(t_1) = m(t_2)\), then \(\pth(t_1) = \pth(t_2)\) from the monotonicity property. Indeed, supposing for instance that \(t_1 \leq t_2\), and defining the element \(\theta \coloneqq \pth(t_2) - \pth(t_1) \) we see that on the one hand \(\int_{\Omega} \theta = 0\), on the other hand, \(\theta \geq 0\), so \(\theta = 0\) and thus \(\pth(t_1) = \pth(t_2)\).

  Now, define \(\Masses \coloneqq m([0,1]) = \setc{s\in[0,1]}{\exists t\in[0,1]\, m(t) = s}\).
  Pick \(s \in [0,1]\).

  If \(s \in \Masses\)  we define \(\psi(s) \coloneqq \pth(t)\) where \(m(t) = s\) (and this does not depend on which \(t\) fulfills \(m(t) = s\) from what we said above).
  We remark that \(\int_{\Omega}\psi(s) = \int_{\Omega}\pth(t) = m(t) = s\).

  Now suppose that \(s \not\in \Masses\).
  Define \(s_1 \coloneqq \sup(\Masses \cap [0,s])\) and \(s_2 \coloneqq \inf(\Masses\cap[s,1])\) (neither set are empty since \(0\in \Masses\) and \(1 \in \Masses\)).
  Since \(s_1\in \Masses\)  and \(s_2\in \Masses\), there are \(t_1\in[0,1]\) and \(t_2\in[0,1]\) such that \(m(t_1) = s_1\) and \(m(t_2) = s_2\).
  Finally define \(\psi(s) \coloneqq \pth(t_1) + (s - s_1)\frac{\pth(t_2) - \pth(t_1)}{s_2 - s_1} \).
  In this case, \(\int_{\Omega}\psi(s) = m(t_1) + (s-s_1) \frac{m(t_2) - m(t_1)}{s_2-s_1} = s\).
  The path \(\psi\) constructed this way is still monotone, and it has the constant speed property,
  so it is an ablation path.
\end{proof}

\section{\(\Lspace\infty\)-optimal Monotonicity Projection}
The algorithm proposed in \autoref{sec:pathopt-algo-details} for optimising monotone paths uses updates that can locally introduce nonmonotonicity in the candidate \(\hat\pth_1\), so that it is necessary to project back onto a monotone path \(\pth_1\). The following routine\footnote{
It is easy to come up with other algorithms for monotonising a (discretised) function. One could simply \emph{sort the array}, but that is not optimal with respect to any of the usual function norms; or clip the derivatives to be nonnegative and then rescale the entire function, but that is not robust against noise pertubations.}
performs such a projection in a way that is optimal in the sense of minimising the \(\Lspace\infty\)-distance\footnote{Note that the optimum is not necessarily unique.}, i.e.,
\[
  \sup_{t}\bigl|\pth_1(t,\vcr) - \hat\pth_1(t,\vcr)\bigr|
    \leq \sup_{t}\bigl|\vartheta(t,\vcr) - \hat\pth_1(t,\vcr)\bigr|
\]
for all \(\vcr\in\Omega\) and any other monotone path \(\vartheta\).

The algorithm works separately for each \(\vcr\), i.e., we express it as operating simply on continuous functions \(p: [0,1]\to\mathbb{R}\).
\begin{algorithm}
  \caption{Make a function \([0,1] \to \RR\) nondecreasing}
  \begin{algorithmic}
    
  \State{\(\cup_{i} [l_i,r_i] \gets \setc{t\in[0,1]}{p'(t) \leq 0}\)}\Comment{Union of intervals where \(p\) decreases}
  \For{\(i\)}
  \State{\(m_i \gets \tfrac{p(l_i)+p(r_i)}{2}\)}
  \State{\(l_i \gets \max \setc{t \in [r_{i-1},l_i]}{p(t) \leq m_i}\)}
    \State{\(r_i \gets \min \setc{t \in [r_i, l_{i+1}]}{p(t) \geq m_i}\)}
  \EndFor
  \For{\(i,j\)}
  \If{\([l_i,r_i] \cap [l_j,r_j] \neq \emptyset\)}
    \State if \(m_j<m_i\), merge the intervals and recompute \(m\) as the new center
    \EndIf
  \EndFor \\
  \Return \(t \mapsto \begin{cases} p(t) & \text{if \(t \not\in \cup_i[l_i,r_i]\)} \\
    m_i & \text{if \(t\in [l_i,r_i]\)}
  \end{cases}\) 
\end{algorithmic}
 \label{monotonicityEnforcement}
\end{algorithm}
The final step effectively \emph{flattens out}, in a minimal way, any region in which the function was decreasing.
\begin{figure}[htb]
  \centering
 \parbox{.49\textwidth}{
   \begin{subfigure}{\linewidth}
     \includegraphics[width=\textwidth]{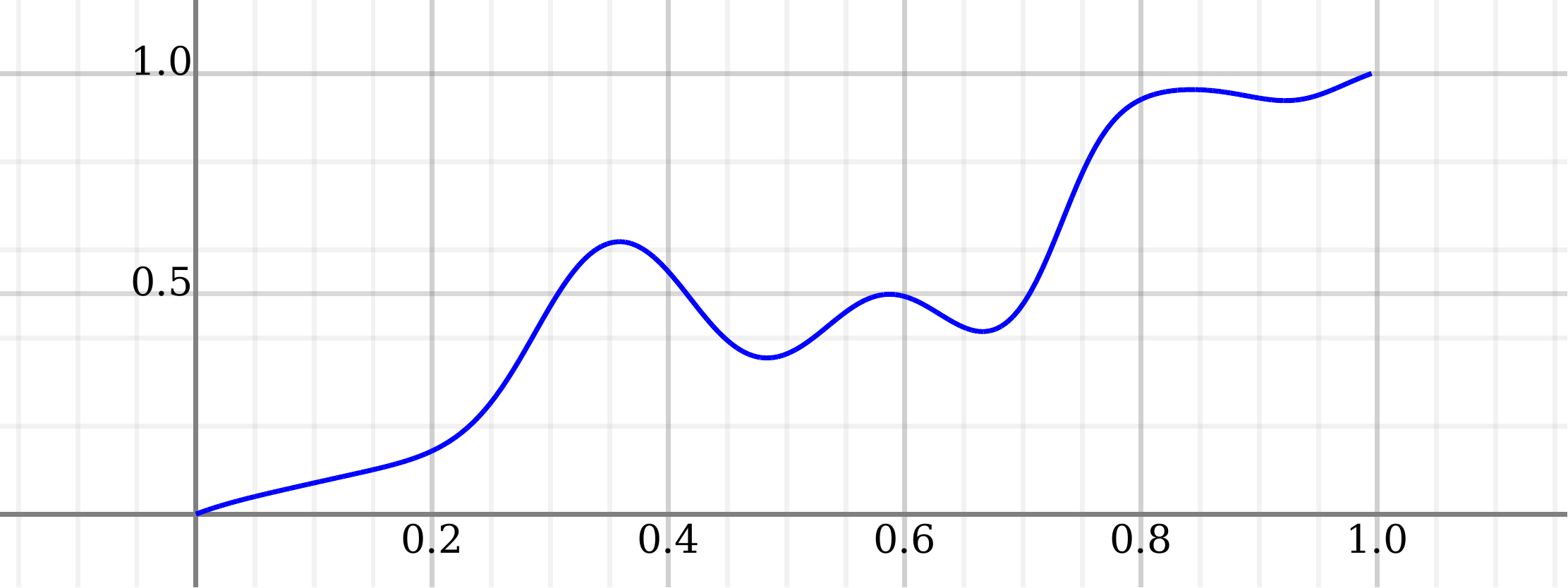}
     \caption{Original}
     \label{monAlgEx-original}
   \end{subfigure}
   \addtocounter{subfigure}{3}
   \begin{subfigure}{\linewidth}
     \vspace{0.15\linewidth}
     \centering
     \includegraphics[width=\textwidth]{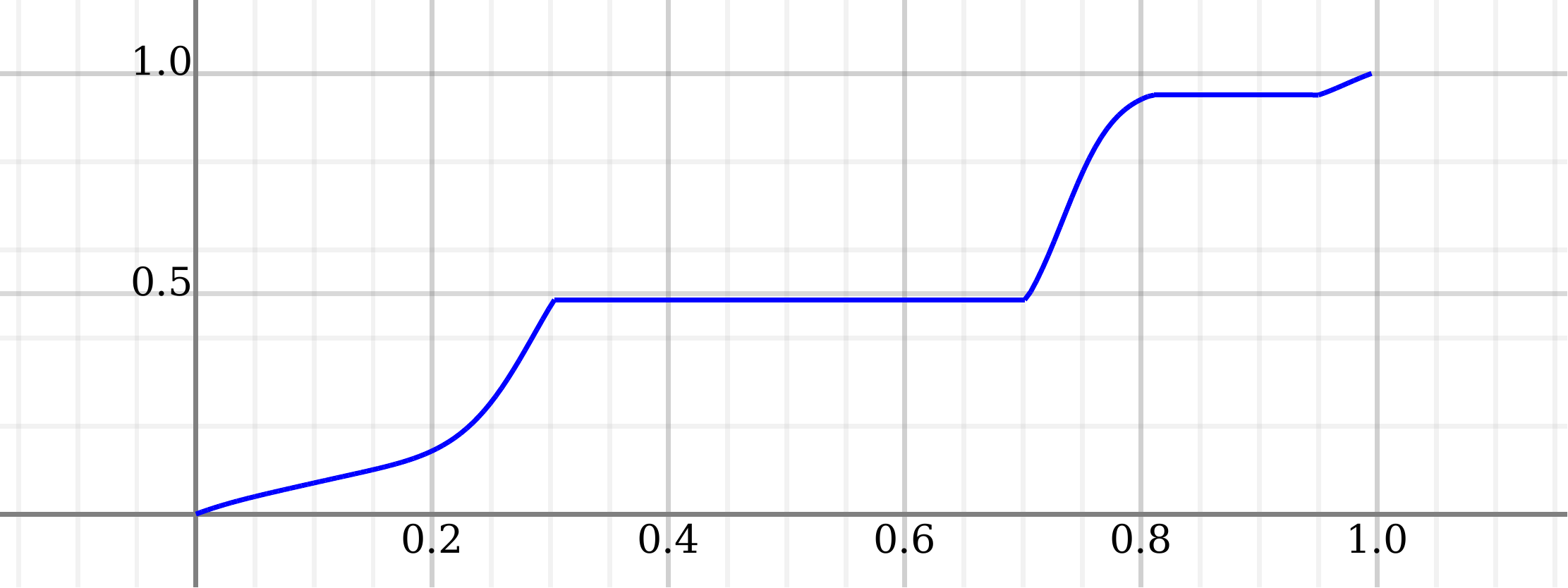}
     \caption{Monotone projection}
     \label{monAlgEx-result}
   \end{subfigure}
  }
   \addtocounter{subfigure}{-4}
 \parbox{.49\textwidth}{
   \begin{subfigure}{\linewidth}
     \centering
     \includegraphics[width=\textwidth]{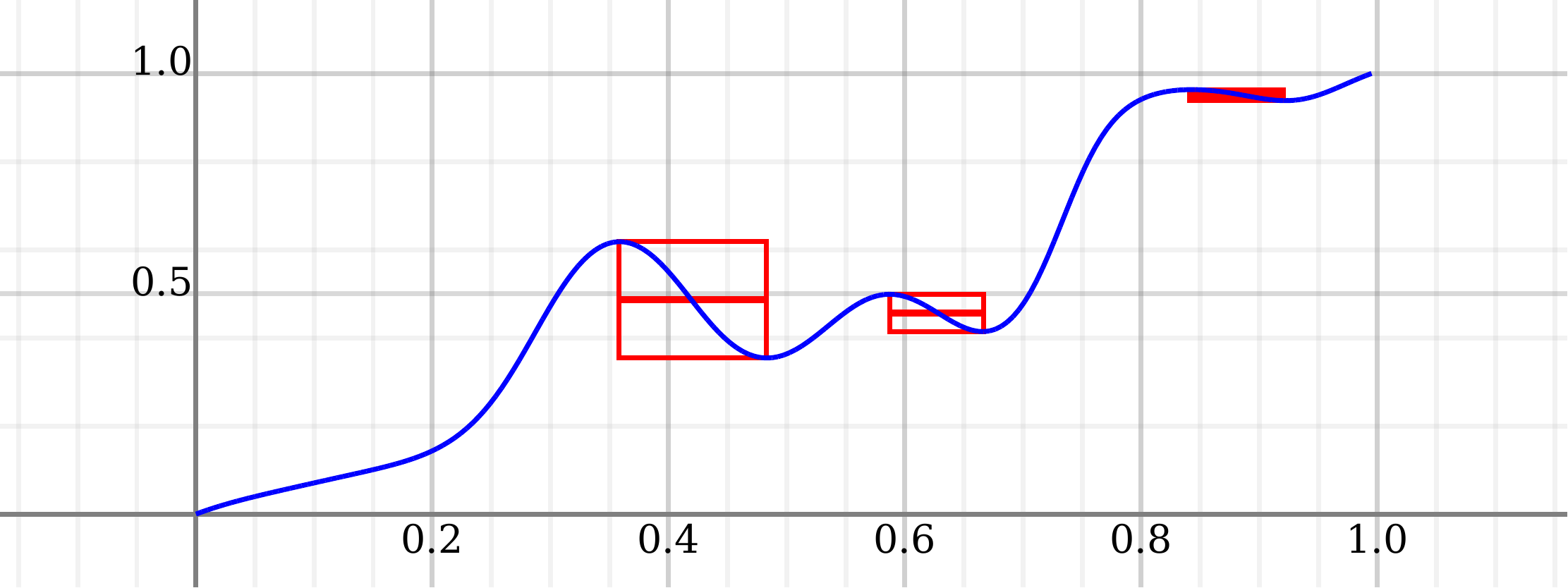}
     \caption{Decreasing intervals}
     \label{monAlgEx-decrIvs}
   \end{subfigure}
   \begin{subfigure}{\linewidth}
     \centering
     \includegraphics[width=\textwidth]{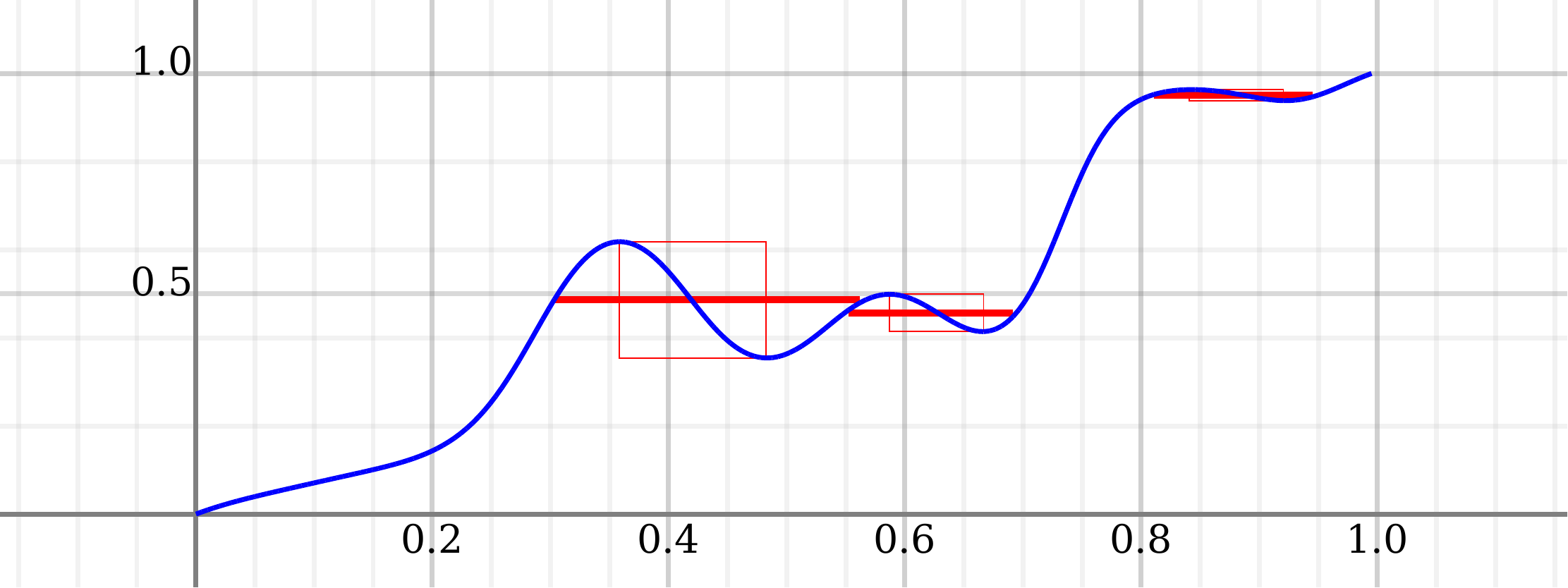}
     \caption{Grown intervals}
     \label{monAlgEx-grownIvs}
   \end{subfigure}
   \begin{subfigure}{\linewidth}
     \centering
     \includegraphics[width=\textwidth]{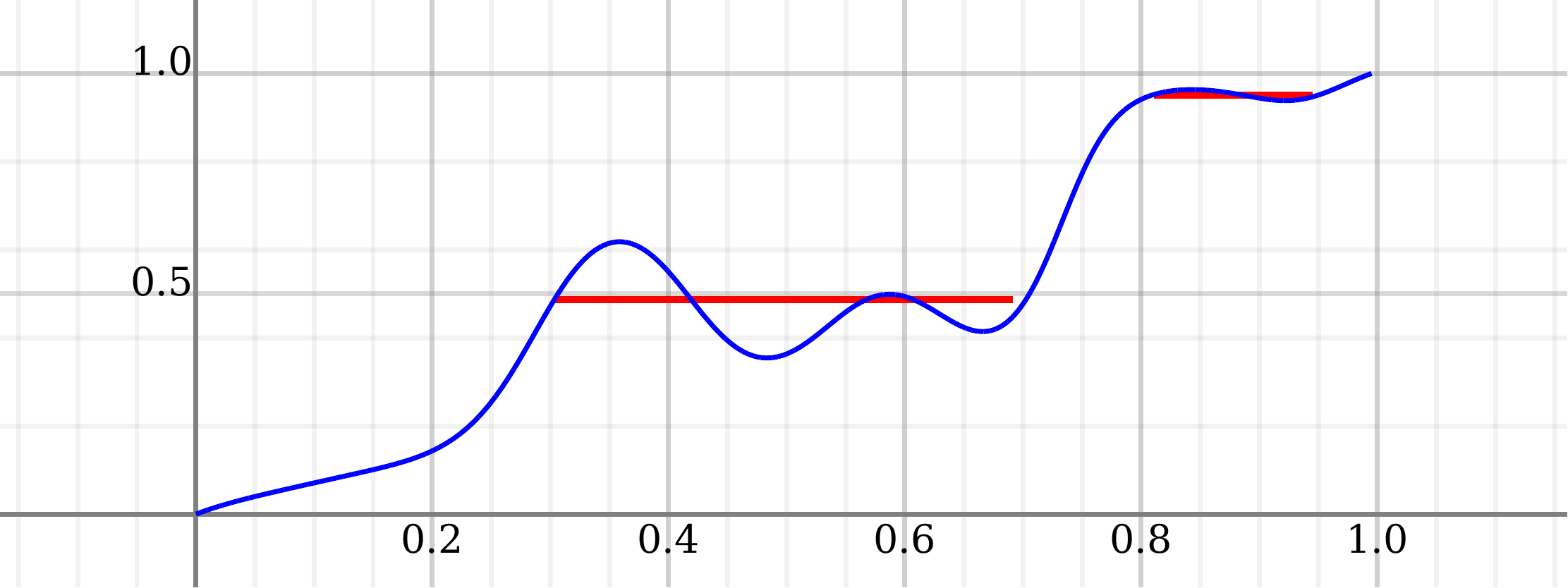}
     \caption{Merged intervals}
     \label{monAlgEx-mergedIvs}
   \end{subfigure}
  }
\caption{%
Example view of the monotonisation algorithm in practice. (\subref{monAlgEx-original}) contains decreasing intervals, which have been localised in (\subref{monAlgEx-decrIvs}). For each interval, the centerline is then extended to meet the original path non-decreasingly (\subref{monAlgEx-grownIvs}). In some cases, this will cause intervals overlapping; in this case merge them to a single interval and re-grow from the corresponding centerline (\subref{monAlgEx-mergedIvs}). Finally, replace the path on the intervals with their centerline (\subref{monAlgEx-result}).%
}
\label{monotonisationexample}
\end{figure}

In practice, this algorithm is executed not on continuous functions but on a PCM-discretised representation; this changes nothing about the algorithm except that instead as real numbers, \(l,r\) and \(t\) are represented by integral indices.
\label{sec:monotonicityprojection}

\section{Path Optimisation Algorithm}\label{sec:pathopt-algo-details}
As said in \autoref{sec:optimisation-idea}, our optimisation algorithm is essentially gradient descent of a path \(\pth\): it repeatedly seeks the direction within the space of all paths that (first ignoring the monotonicity constraint) would affect the largest increase to \(\Score(\pth)\) as per \autoref{sec:score}, for any of the defined score functions. \autoref{algo:projGradDesc} shows the details of how this is done in presence of our constraints. In case of \(\ScoreBndStr\), the state \(\pth\) is understood to consist of the two paths \(\pthRet\) and \(\pthDiss\).
\begin{algorithm}
  \caption{Projected Gradient Descent}
  \label{algo:projGradDesc}
\begin{algorithmic}[1]
  \State \(\pth \gets ((t,\mathbf{r})\mapsto t)\) \Comment Start with linear-interpolation path
  \While{\(\pth\) is not sufficiently saturated}
    \For{\(t\) in \([0,1]\)}
      \State \(x_{\pth,t} := (1-\pth(t))\: \curIm + \pth(t)\: \basLn \)
      \State compute \(F(x_{\pth,t})\) with gradient
               \(\mathbf{g}:=\nabla F(x_{\pth,t})\)
      \State let \(\hat{\mathbf{g}} := \mathbf{g} - \int_\Omega \mathbf{g}\)
              \Comment{ensure \(\hat{\mathbf{g}}\) does not affect mass of \(\pth(t)\)}
      \State update \(\pth(t,\vcr) \gets
              \pth(t,\vcr)
              - \gamma\:\left\langle
                  \hat{\mathbf{g}}(\mathbf{r}) \mid| \basLn-\curIm \right\rangle\)
              , for \(\vcr\) in \(\Omega\)
      \State (optional) apply a regularisation filter to \(\pth(t)\)
    \EndFor
    \State (optional) adjust learning rate \(\gamma\) according to size of the actual step performed
    \State (optional) apply saturation to \(\pth\) (\autoref{sec:maskSaturation})
    \State (optional) apply pinching to the paths \(\pthRet,\pthDiss\)
                              (\autoref{sec:boundaryPinching})
    \For{\(\mathbf{r}\) in \(\Omega\)}
       \State re-monotonise \(t \mapsto \pth(t,\vcr)\), using \autoref{monotonicityEnforcement}
    \EndFor
    \State clamp \(\pth(t,\vcr)\) to \([0,1]\) everywhere
    \State re-parametrise \(\pth\), such that \(\int_\Omega \pth(t) = t\) for all \(t\) (using \autoref{sec:constantspeed}) 
  \EndWhile.
\end{algorithmic}
\end{algorithm}

As discussed before, the use of a gradient requires a metric to obtain a vector from the covector-differential, which could be either the implicit \(\ell^2\) metric on the discretised representation (pixels), or a more physical kernel/filter-based metric.
In the present work, we base this on the regularisation filter.

Unlike with the monotonisation condition, the update can easily be made to preserve speed-constness by construction, by projecting for each \(t\) the gradient \(\mathbf{g}\) on the sub-tangent-space of zero change to \(\int_\Omega\pth(t)\), by subtracting the constant function times \(\int_\Omega\mathbf{g}(t)\). Note this requires the measure of \(\Omega\) to be normalised, or else considered at this point.

Then we apply these gradients time-wise as updates to the path, using a scalar product in the channel-space to obtain the best direction for \(\pth\) itself (as opposed to the corresponding image composit \(x_{\pth,t}\)).

The learning rate \(\gamma\) can be chosen in different ways. What we found to work best is to normalise the step size in a \(\Lspace\infty\) sense, such that the strongest-affected pixel in the mask experiences a change of at most 0.7 per step. This is small enough to avoid excessively violating the constraint, but not so small to make the algorithm unnecessarily slow.

\section{Baseline choice}
\label{sec:appendixbaselineChoice}

\begin{figure}[htb]
  \centering
    \includegraphics[width=\textwidth]{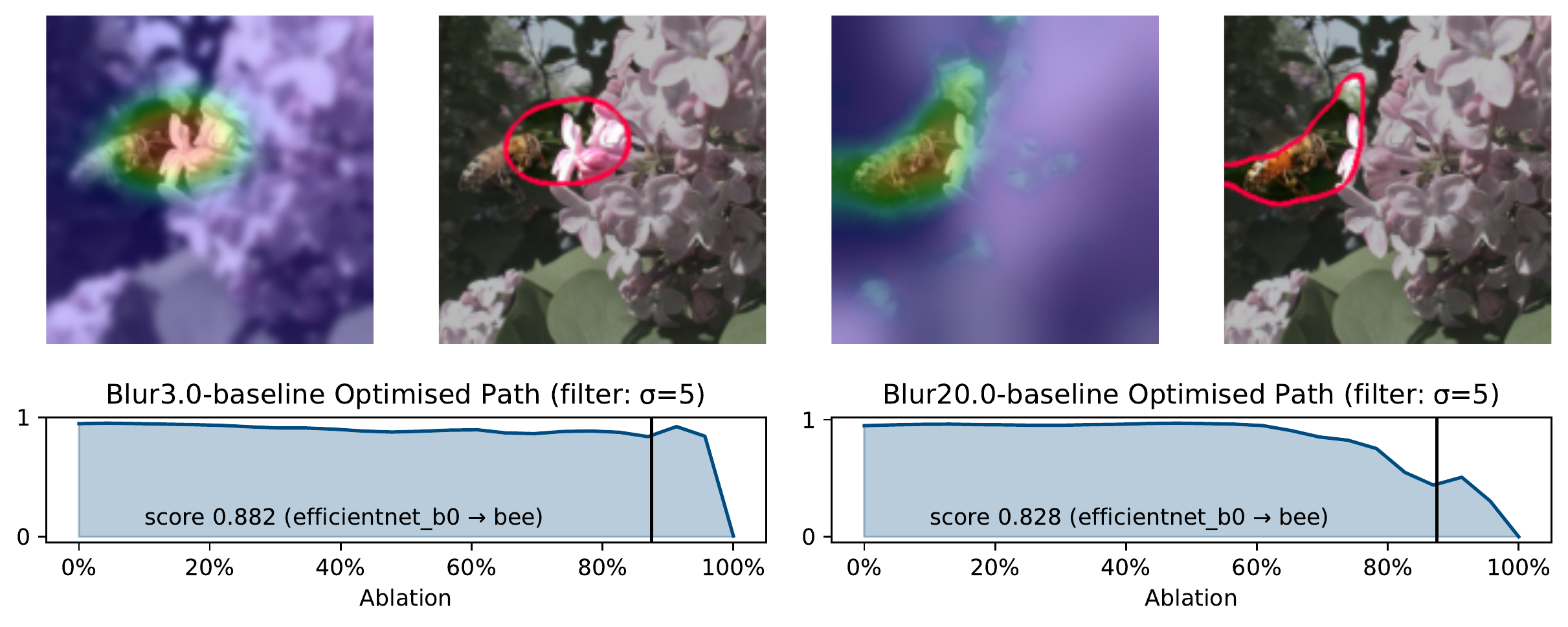}
\caption{%
An example of paths obtained with different-size blur baselines.%
}
\label{fig:ablationcompare-baselineBlurrins}
\end{figure}

The baseline image is prominently present in the input for much of the ablation path, and it is therefore evident that it will have a significant impact on the saliency. In line with previous work, we opted for a blurred baseline for the examples in the main paper, but even then there is still considerable freedom in the choice of blurring filter. \autoref{fig:ablationcompare-baselineBlurrins} shows two examples, where the result is not fundamentally, but still notably different.

\end{document}